\documentclass[10pt,twocolumn,letterpaper]{article}

\usepackage{cvpr}
\usepackage{times}
\usepackage{epsfig}
\usepackage{amsmath}
\usepackage{paralist}
\usepackage{amssymb}
\usepackage{float}
\usepackage[hang]{subfigure}
\usepackage{graphicx, array, booktabs}
\usepackage[export]{adjustbox}
\usepackage{multirow}
\usepackage{tabularx}
\usepackage{algorithm}
\usepackage{algorithmicx}
\usepackage{enumerate}
\usepackage{multicol}

\usepackage[pagebackref=true,breaklinks=true,letterpaper=true,colorlinks,bookmarks=false]{hyperref}

\cvprfinalcopy 

\def\cvprPaperID{2298} 

\ifcvprfinal\fi
\begin{document}

\title{Noise-Aware Unsupervised Deep Lidar-Stereo Fusion}


\author{\thanks{These authors contributed equally in this work.} Xuelian Cheng$^{1,2}$, $^{\scriptsize*}$Yiran Zhong$^{2,4,5}$, Yuchao Dai$^{1}$, Pan Ji$^{3}$, Hongdong Li$^{2,4}$\\
$^{1}$Northwestern Polytechnical University
$^{2}$Australian National University\\
$^{3}$NEC Laboratories America, $^{4}$ACRV, $^{5}$Data61 CSIRO \\
}
\maketitle
\thispagestyle{empty}

\begin{abstract}

In this paper, we present LidarStereoNet, the first unsupervised Lidar-stereo fusion network, which can be trained in an end-to-end manner without the need of ground truth depth maps. By introducing a novel ``Feedback Loop'' to connect the network input with output, LidarStereoNet could tackle both noisy Lidar points and misalignment between sensors that have been ignored in existing Lidar-stereo fusion studies. Besides, we propose to incorporate a piecewise planar model into network learning to further constrain depths to conform to the underlying 3D geometry. Extensive quantitative and qualitative evaluations on both real and synthetic datasets demonstrate the superiority of our method, which outperforms state-of-the-art stereo matching, depth completion and Lidar-Stereo fusion approaches significantly.

\end{abstract}

\section{Introduction}
Accurately perceiving surrounding 3D information from passive and active sensors is crucial for numerous applications such as localization and mapping \cite{kelly2011visual}, autonomous driving \cite{levinson2011towards}, obstacle detection and avoidance \cite{oniga2010processing}, and 3D reconstruction \cite{geiger2011stereoscan, zhang2015meshstereo}. However, each kind of sensors alone suffers from its inherent drawbacks. Stereo cameras are well-known for suffering from computational complexities and their incompetence in dealing with textureless/repetitive areas and occlusion regions \cite{scharstein2002taxonomy}, while Lidar sensors often provide accurate but relatively sparse depth measurements \cite{deems2013lidar}. 

\begin{figure}[h]
    \centering
    \tabcolsep=0.05cm
    \begin{tabular}{c c}
    \includegraphics[width=0.485\linewidth]{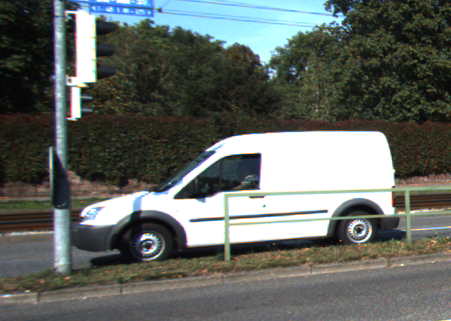} &
    \includegraphics[width=0.485\linewidth]{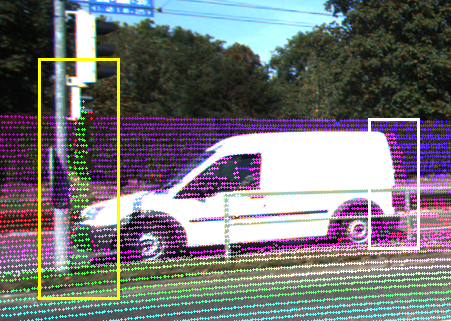} \\
    \small{Input} & \small{Lidar} \\
    \includegraphics[width=0.485\linewidth]{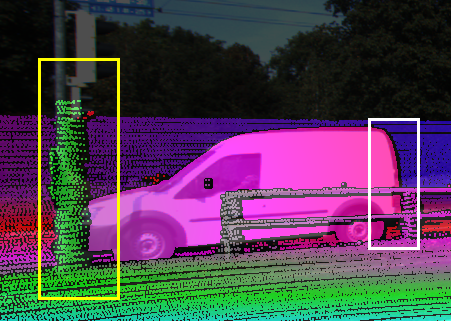}   &
    \includegraphics[width=0.485\linewidth]{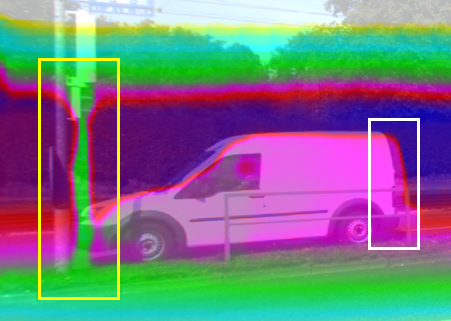} \\
    \small{GT} & \small{S2D \cite{Ma2018SparseToDense}}   \\
    \includegraphics[width=0.485\linewidth]{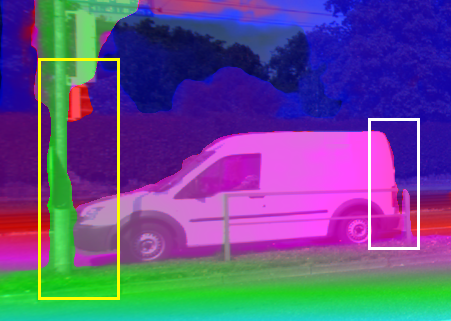}   &
    \includegraphics[width=0.485\linewidth]{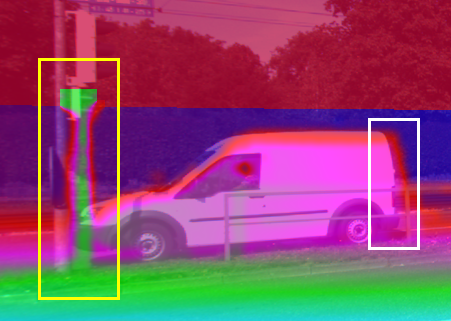} \\
    \small{Ours} & \small{SINet \cite{Uhrig2017THREEDV}} \\
    \end{tabular}
    \caption{\small \textbf{Results on KITTI 2015.} We highlight the displacement error of Lidar points with bounding boxes. Lidar points are dilated for better visualization and we overlay our disparity maps to the colour images for illustration. Note the Lidar points for the foreground car and utility pole have been aligned to the background. Our method successfully recovers accurate disparities on tiny and moving objects while the other methods are misled by drifted and noisy Lidar points.}
    \label{fig:lidar_error}
    \vspace{-3mm}
\end{figure}

Therefore, it is highly desired to fuse measurements from Lidar and stereo cameras to achieve high-precision depth perception by exploiting their complementary properties. 
However, it is a non-trivial task as accurate Stereo-Lidar fusion requires a proper registration between Lidar and stereo images and noise-free Lidar points. Existing methods are not satisfactory due to the following drawbacks:
\begin{compactitem}
    \item Existing deep neural network based Lidar-Stereo fusion studies \cite{Badino2011, Maddern2016, Park2018} strongly depend on the availability of large-scale ground truth depth maps, and thus their performance is fundamentally limited by their generalization ability to real-world applications.
    \item Due to rolling-shutter effects of Lidar and other calibration imperfections, a direct registration will introduce significant alignment errors between Lidar and stereo depth. Furthermore, existing methods tend to assume the Lidar measurements are noise-free \cite{Ma2018SparseToDense,Uhrig2017THREEDV}. However, as illustrated in Fig.~\ref{fig:lidar_error}, the misalignment and noisy Lidar measurements cause significant defects in Stereo-Lidar fusion.
\end{compactitem}
In this paper, we tackle the above challenges and propose a novel framework ``LidarStereoNet'' for accurate Stereo-Lidar fusion, which can be trained in an end-to-end unsupervised learning manner. Our framework is noise-aware in the sense that it explicitly handles misalignment and noise in Lidar measurements. 

Firstly, we propose to exploit photometric consistency between stereo images, and depth consistency between stereo cameras and Lidar to build an unsupervised training loss, thus removing the need of ground truth depth/disparity maps. It enables a strong generalization ability of our framework to various real-world applications. 

Secondly, to alleviate noisy Lidar measurements and slight misalignment between stereo cameras and Lidar, we present a novel training strategy that gradually removes these noisy points during the training process automatically. Furthermore, we have also presented a novel structural loss (named plane fitting loss) to handle the inaccurate Lidar measurements and stereo matching. 

Under our problem setting, we make no assumption on the inputs such as the pattern/number of Lidar measurements, the probability distribution of Lidar points or stereo disparities. Our network allows the sparsity of input Lidar points to be varied, and can even handle an extreme case when the Lidar sensor is completely unavailable. 

Experimental results on different datasets demonstrate that our method is able to recover highly accurate depth maps through Lidar-Stereo fusion. It outperforms existing stereo matching methods, depth completion methods and Lidar-Stereo fusion methods with a large margin (at least twice better than previous ones). To the best of our knowledge, there is no deep learning based method available that can achieve this goal under our problem setting.
\section{Related Work}

\paragraph{Stereo Matching}
Deep convolutional neural networks (CNNs) based stereo matching methods have recently achieved great success. Existing supervised deep methods either formulate the task as depth regression \cite{Mayer2016CVPR} or multi-label class classifications \cite{Zbontar2016}. Recently, unsupervised deep stereo matching methods have also been introduced to relief from a large amount of labeled training data. Godard \etal \cite{godard2017unsupervised} proposed to exploit the photometric consistency loss between left images and the warped version of right images, thus forming an unsupervised stereo matching framework. Zhong \etal \cite{Zhong2018ECCV_rnn} presented a stereo matching network for estimating depths from continuous video input. Very recently, Zhang \etal \cite{Zhang2018ECCV} extended the self-supervised stereo network \cite{zhong2017self} from passive stereo cameras to active stereo scenarios. Even though stereo matching has been greatly advanced, it still suffers from challenging scenarios such as texture-less and low-lighting conditions. 

\vspace{-3mm}
\paragraph{Depth Completion/Interpolation}
Lidar scanners can provide accurate but sparse and incomplete 3D measurements. Therefore, there is a highly desired requirement in increasing the density of Lidar scans, which is crucial for applications such as self-driving cars. Uhrig \etal \cite{Uhrig2017THREEDV} proposed a masked sparse convolution layer to handle sparse and irregular Lidar inputs. Chodosh \etal \cite{chodosh2018deep} utilized compressed sensing to approach the sparsity problem for scene depth completion. With the guidance of corresponding color images, Ma \etal \cite{Ma2018SparseToDense} extended the up-projection blocks proposed by \cite{laina2016deeper} as decoding layers to achieve full depth reconstruction. Jaritz \etal \cite{jaritz2018sparse} handled sparse inputs of various densities without any additional mask input.

\vspace{-3mm}
\paragraph{Lidar-Stereo Fusion}
Existing studies mainly focus on fusing stereo and time-of-flight (ToF) cameras for indoor scenarios \cite{ELOMARI20081,moghadam2008improving, Nickels:113764, Harrison2009, Gandhi2012}, while Lidar-Stereo fusion for outdoor scenes has been seldom approached in the literature. Badino \etal \cite{Badino2011} used Lidar measurements to reduce the searching space for stereo matching and provided predefined paths for dynamic programming. Later on, Maddern \etal \cite{Maddern2016} proposed a probabilistic model to fuse Lidar and disparities by combining prior from each sensor. However, their performance degrades significantly when the Lidar information is missing. To tackle this issue, instead of using a manually selected probabilistic model, Park \etal \cite{Park2018} utilized CNNs to learn such a model, which takes two disparities as input: one from the interpolated Lidar and the other from semi-global matching \cite{Hirschmuller2008}. Compared with those supervised approaches, our unsupervised method can be end-to-end trained using stereo pairs and sparse Lidar points without using external stereo matching algorithms. 

\section{Lidar-Stereo Fusion}
In this section, we formulate Lidar-Stereo fusion as an unsupervised learning problem and present our main ideas in dealing with the inherent challenges encountered by existing methods, \ie, noise in Lidar measurements.

\subsection{Problem Statement}
Lidar-Stereo fusion aims at recovering a dense and accurate depth/disparity map from sparse Lidar measurements $S\in\mathbb{R}^{n\times3}$ and a pair of stereo images $I_l$, $I_r$. We assume the Lidar and stereo camera have been calibrated with extrinsic matrix $T$ and the stereo camera itself is calibrated with intrinsic matrices $K_l$, $K_r$ and projection matrices $P_l$, $P_r$. We can then project sparse Lidar points $S$ onto the image plane of $I_l$ by $d^s_l = P_l T S$. Since disparity is used in the stereo matching task, we convert the projected depth $d^s_l$ to disparity $D^s_l$ using $D = Bf/d$, where $B$ is the baseline between the stereo camera pair and $f$ is the focal length. The same process is applied to the right image as well. Mathematically this problem can be defined as:
\begin{equation}
    (\widehat{D}_l, \widehat{D}_r) = \mathcal{F}(I_l,I_r,D^s_l,D^s_r;\Theta),
\end{equation}
where $\mathcal{F}$ is the learned Lidar-Stereo fusion model (a deep network in our paper) parameterized by $\Theta$, $\widehat{D}_l, \widehat{D}_r$ are the fusion outputs defined on the left and right coordinates.

Under our problem setting, we do not make any assumption on the Lidar points' configuration (\eg, the number or the pattern) or error distribution of Lidar points and stereo disparities. Removing all these restrictions makes our method more generic and wider applicability.

\subsection{Dealing with Noise}
\begin{figure}[t]
\centering
   \includegraphics[width=1.0\linewidth]{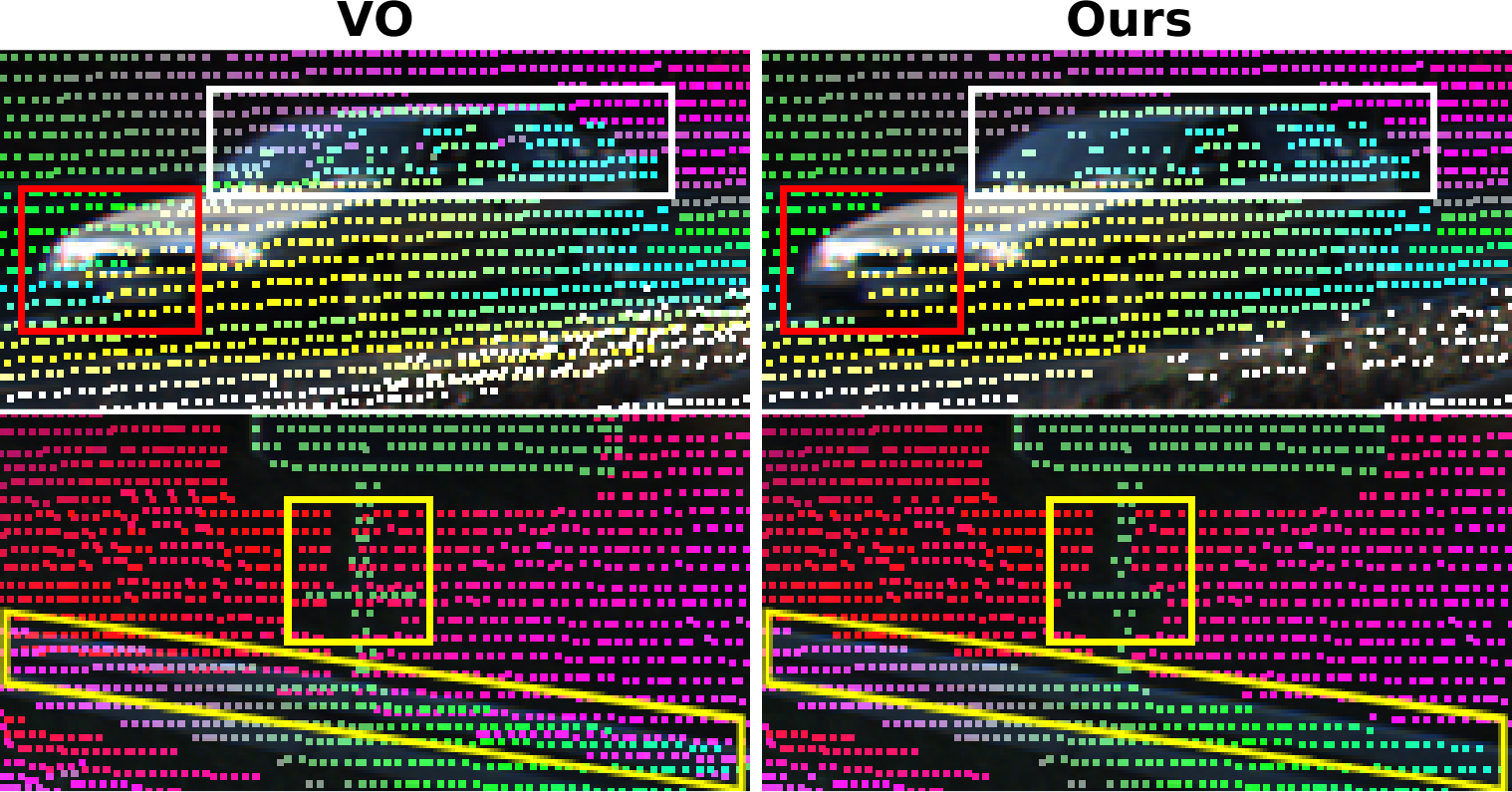}
   \caption{\textbf{KITTI VO Lidar points and our cleaned Lidar points.} Erroneous Lidar points on transparent/relective areas and tiny object surface have been successfully removed.}
   \label{fig:lidar_error2}
\end{figure}

In Lidar-Stereo fusion, existing methods usually assume the Lidar points are noise free and the alignment between Lidar and stereo images is perfect. We argue that even for dedicated systems such as the KITTI dataset, the Lidar points are never perfect and the alignment cannot be consistently accurate, \cf Fig.~\ref{fig:lidar_error}. The errors in Lidar scanning are inevitable for two reasons:
(\textbf{1}) Even for well calibrated Lidar-stereo systems, \emph{e.g.}, KITTI, and after eliminating the rolling shutter effect in Lidar scans by compensating the ego-motion, Lidar errors still persist even for stationary scenes, as shown in Fig.~\ref{fig:lidar_error2}. According to the readme file in the KITTI VO dataset, the rolling shutter effect has already been removed in Lidar scans. However, we still find Lidar errors on transparent (white box) and reflective (red box) surfaces. Also, due to the displacement between Lidar and cameras, the Lidar can see through tiny objects as shown in the yellow box. 
(\textbf{2}) It is hard to perform motion-compensation on dynamic scenes, thus the rolling shutter effect will persist for moving objects.

It is possible to eliminate these errors by manually inserting 3D models and other post-processing steps \cite{Menze2015ISA}. However, lots of human efforts will be involved. Our method can automatically deal with these Lidar errors without the need of human power. Hence the problem we are tackling (``Noise-Aware Lidar-Stereo Fusion'') is not a simple ``system-level'' problem that can be solved through ``engineering registration''. 

It is known that the ability of deep CNNs to overfit or memorize the corrupted labels can lead to poor generalization performance. Therefore, we aim to deal with the noise in Lidar measurements properly to train deep CNNs.

Robust functions such the $\ell_1$ norm, Huber function or the truncated $\ell_2$ norm are natural choices in dealing with noisy measurements. However, these functions will not eliminate the effects caused by noises but only suppress them. Further, these errors also exist in the input. Automatically correct/ignore these erroneous points creates an extra difficulty for the network. To this end, we introduce a \emph{feedback loop} in our network to allow the input also to depend on the output of the network. In this way, the input Lidar points can be cleaned before being fed into the network. 

\begin{figure}[t]
    \centering
    \includegraphics[width=0.9\linewidth]{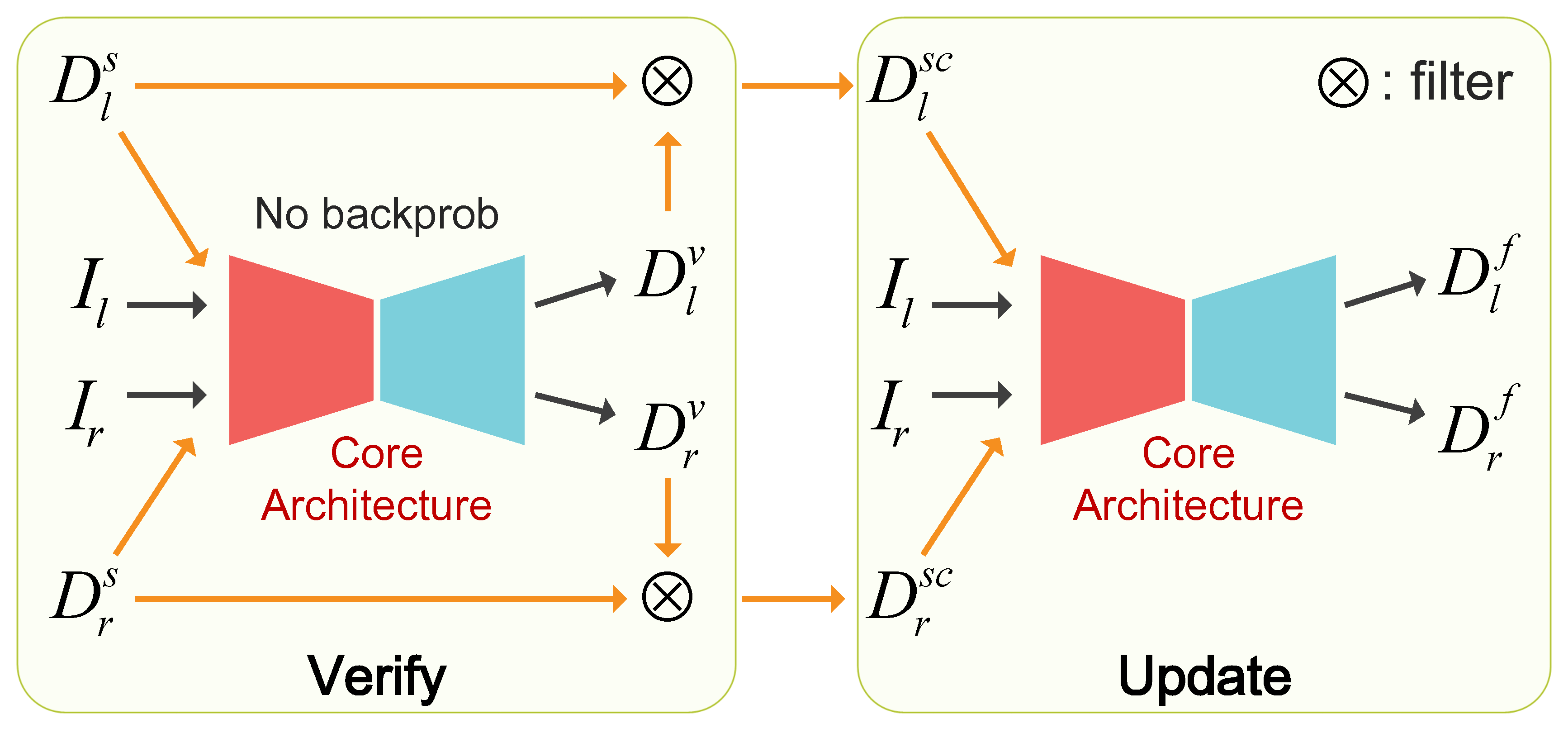}
    \caption{\textbf{The feedback loop.} For each iteration, the input stereo pair first computes initial disparities to filter errors in sparse Lidar points. At this stage, no \emph{backprob} is taken place. So we call it the Verify phase. Then in the Update phase, the Core Architecture takes stereo pairs and cleaned sparse Lidar points as inputs to generate the final disparities. The parameters of the Core Architecture will be updated through \emph{backprob} this time.}
    \label{fig:feedbackloop}
\end{figure}

\vspace{-3mm}
\paragraph{The Feedback Loop}
We propose a novel framework to progressively detect and remove erroneous Lidar points during the training process and generate a highly accurate Lidar-Stereo fusion. Fig.~\ref{fig:feedbackloop} illustrates an unfolded structure of our network design, namely the feedback loop. It consists of two phases: ``Verify'' phase and ``Update'' phase. Each phase shares the same network structure of Core Architecture, and the details will be illustrated in Section \ref{core}. 

\begin{figure*}[t]
    \centering
    \includegraphics[width=1\linewidth]{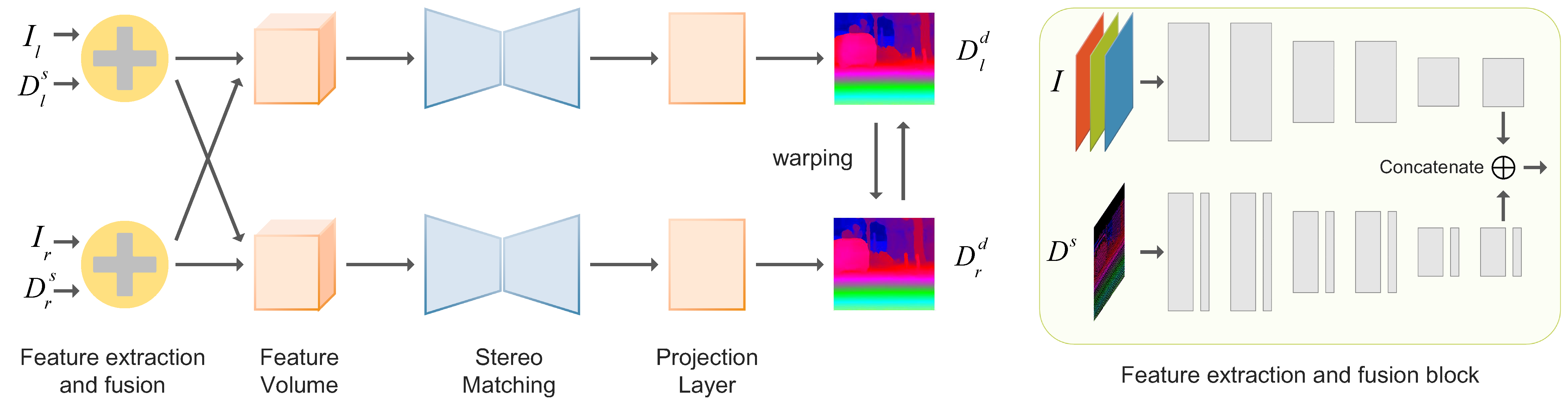}
    \caption{\textbf{Core Architecture of our LidarStereoNet.} It consists of a feature extraction and fusion block, a stack-hourglass type feature matching block and a disparity computing layer. Given a stereo pair $I_l, I_r$ and corresponding projected Lidar points $D_l^s, D_r^s$, the feature extraction block produces feature maps separately for images and Lidar points. The feature maps are then concatenated to form final input features which are aggregated to form a feature volume. The feature matching block learns the cost of feature-volume. Then we use the disparity computing layer to obtain disparity estimation. Details of the feature extraction and fusion block is illustrated on the right.}
    \label{fig:core_structure}
\end{figure*}

In the Verify phase, the network takes stereo image pairs ($I_l$,$I_r$) and noisy Lidar disparities ($D_l^s$, $D_r^s$) as input, and generates two disparity maps ($D^v_l$,$D^v_r$). No back-propagation takes place in this phase. We then compare ($D^v_l$,$D^v_r$) and ($D_l^s$,$D_r^s$) and retain the sparse Lidar points ($D_l^{sc}$,$D_r^{sc}$) that are consistent in both stereo matching and Lidar measurements. In the Update phase, the network takes both stereo pairs ($I_l$,$I_r$) and cleaned sparse Lidar points ($D_l^{sc}$,$D_r^{sc}$) as the inputs to recover dense disparity maps ($D_l^{f}$,$D_r^{f}$). All loss functions are evaluated on the final disparity outputs ($D_l^{f}$,$D_r^{f}$) only. Once the network is trained, we empirically find that there is no performance drop if we directly feed the Core Architecture with noisy Lidar points. Therefore, we remove the feedback loop module and only use the Core Architecture in testing.

Our feedback loop detects erroneous Lidar points by measuring the consistency between Lidar points and stereo matching. Lidar and stereo matching are active and passive depth acquisition techniques. Hence, it is less likely that they would make the same errors. It may also filter out some correct Lidar points at the first place but we have image warping loss and other regularization losses to keep the network training on the right track. 

\section{Our Network Design}
In this section, we present our ``LidarStereoNet'' for Lidar-Stereo fusion, which can be learned in an unsupervised end-to-end manner. To remove the need of large-scale training data with ground truth, we propose to exploit the photometric consistency between stereo images, and the depth consistency between stereo cameras and Lidar. This novel network design enables the following benefits: \textbf{1}) A wide generalization ability of our framework in various real-world scenarios; \textbf{2}) Our network design allows the sparsity of input Lidar points to be varied, and can even handle the extreme case when the Lidar sensor is completely unavailable. Furthermore, to alleviate noisy Lidar measurements and the misalignment between Lidar and stereo cameras, we incorporate the ``Feedback Loop'' into the network design to connect the output with the input, which enables the Lidar points to be cleaned before fed into the network.

\subsection{Core Architecture}
\label{core}
We illustrate the detailed structure of the Core architecture of our LidarStereoNet in Fig.~\ref{fig:core_structure}. LidarStereoNet consists of the following blocks: \textbf{1}) Feature extraction and fusion; \textbf{2}) Feature matching and \textbf{3}) Disparity computing. The general information flow of our network is similar to \cite{zhong2017self} but has some crucial modifications in the feature extraction and fusion block. In view of different characteristics between dense colour images and sparse disparity maps, we leverage different convolution layers to extract features from each of them. For colour images, we use the same feature extraction block from \cite{chang2018pyramid} while for sparse Lidar inputs, the sparse invariant convolution layer \cite{Uhrig2017THREEDV} is used. The final feature maps are produced by concatenating stereo image features and Lidar features. Feature maps from left and right branches are concatenated to form a 4D feature volume with a maximum disparity range of 192. Then feature matching is processed through an hourglass structure of 3D convolutions to compute matching cost at each disparity level. Similar to \cite{kendall2017end}, we use the soft-argmin operation to produce a 2D disparity map from the cost volume.  

\vspace{-3mm}
\paragraph{Dealing with dense and sparse inputs}
To extract features from sparse Lidar points, Uhrig \etal \cite{Uhrig2017THREEDV} proposed a sparsity invariant CNN after observing the failure of conventional convolutions. However, Ma \etal \cite{Ma2018SparseToDense} and Jaritz \etal \cite{jaritz2018sparse} argued that using a standard CNN with special training strategies can achieve better performance and also handle varying input densities. We compared both approaches and realized that standard CNNs can handle sparse inputs and even get better performance but they request much deeper network (ResNet38 encoded VS 5 Convolutional layers) with 500 times more trainable parameters ($13675.25K$ VS $25.87K$). Using such a ``deep'' network as a feature extractor will make our network not feasible for end-to-end training and hard to converge. 

In our network, we choose sparsity invariant convolutional layers \cite{Uhrig2017THREEDV} to assemble our Lidar feature extractor which can handle varying Lidar points distribution elegantly. It consists of 5 sparse convolutional layers with a stride of 1. Each convolution has an output channel of 16 and is followed by a ReLU activation function. We attached a plain convolution with a stride of 4 to generate the final 16 channels Lidar features in order to make sure the Lidar features compatible with the image features.

\subsection{Loss Function}
Our loss function consists of two data terms and two regularization terms. For data terms, we directly choose the image warping error $\mathcal{L}_w$ as a dense supervision for every pixel and discrepancy on filtered sparse Lidar points $\mathcal{L}_l$. For regularization terms, we use colour weighted smoothness term $\mathcal{L}_p$ and our novel slanted plane fitting loss $\mathcal{L}_p$. Our overall loss function is a weighted sum of the above loss terms:
\begin{equation}
    \mathcal{L} = \mathcal{L}_l + \mu_1\mathcal{L}_w + \mu_2\mathcal{L}_s + \mu_3\mathcal{L}_p,
    \label{eq:all_function}
\end{equation}
we empirically set $\mu_1 = 1,\mu_2 =0.001,\mu_3 = 0.01$. 

\subsubsection{Image Warping Loss}
We assume photometric consistency between stereo pairs such that corresponding points between each pair should have similar appearance. However, in some cases, this assumption does not hold. Hence, we also compare the difference between small patches' Census transform as it is robust for photometric changes. Our image warping loss is defined as follow:
{
\begin{equation}
\mathcal{L}_{w} = \mathcal{L}_{i}+\lambda_{1}\mathcal{L}_{c}+ \lambda_{2}\mathcal{L}_{g},
\end{equation}
}
where $\mathcal{L}_{i}$ stands for photometric loss, $\mathcal{L}_{c}$ represents Census loss and $\mathcal{L}_{g}$ is the image gradient loss. We set $\lambda_{1} = 0.1,\lambda_{2} = 1$ to balance different terms.

The photometric loss is defined as the difference between the observed left (right) image and the warped left (right) image, where we have weighted each term with the observed pixels to account for the occlusion:
{
\begin{equation}
\small
\mathcal{L}_{i} = \left[\sum_{i,j} \varphi\left(\hat{I}(i,j)-I(i,j)\right)\cdot O(i,j)\right]/ \sum_{i,j} O(i,j), 
\end{equation}  
}
where $\varphi(s) = \sqrt{s^{2} + 0.001^{2}}$ and the occlusion mask $O$ is computed through left-right consistency check. 

To further improve the robustness in evaluating the image warping error, we used the Census transformation to measure the difference:
\begin{equation}
\small
\begin{aligned}
\mathcal{L}_{c} = \left[\sum_{i,j} \varphi\left(\hat{C}(i,j)-C(i,j)\right)\cdot O(i,j)\right]/ \sum_{i,j} O(i,j).
\end{aligned}
\end{equation}

Lastly, we have also used the difference between image gradients as an error metric:
\begin{equation}
\small
\mathcal{L}_{g} = \left[\sum_{i,j} \varphi\left(\nabla \hat{I}(i,j)-\nabla I(i,j)\right)\cdot O(i,j)\right]/ \sum_{i,j} O(i,j). 
\end{equation} 

\subsubsection{Lidar Loss}
The cleaned sparse Lidar points after our feedback verification can also be used as a sparse supervision for generating disparities. We leverage the truncated $\ell_2$ function to handle noises and errors in these sparse Lidar measurements,
\begin{equation}
\mathcal{L}_{l} = ||M(\widehat{D} - D^{sc})||_{\tau},
\end{equation}
where $M$ is the mask computed in the Verify phase. The truncated $\ell_2$ fuction is defined as:
\begin{equation}
||\cdot||_\tau = \left\{
             \begin{array}{lc}
             0.5x^2, &  |x| < \epsilon\\
             0.5\epsilon^2, & \mathrm{otherwise}.  
             \end{array}
\right.
\end{equation}

\subsubsection{Smoothness Loss}
The smoothness term in the loss function is defined as:
\begin{equation}
\mathcal{L}_{s} = \sum\left(e^{-\alpha_1\left|\nabla I\right|}\left|\nabla d\right|+e^{-\alpha_2\left|\nabla^2 I\right|}\left|\nabla^2 d\right|\right)/N,
\end{equation}
where $\alpha_1 = 0.5$ and $\alpha_2 = 0.5$. Note that previous studies \cite{godard2017unsupervised,zhong2017self} often neglect the weights $\alpha_1, \alpha_2$, which actually play a crucial role in colour weighted smoothness term. 

\subsubsection{Plane Fitting Loss}
We also introduce a slanted plane model into deep learning frameworks to enforce structural constraint. This model has been commonly used in conventional Conditional Random Field (CRF) based stereo matching/optical flow algorithms. It assumes that all pixels within a superpixel lie on a 3D plane. By leveraging this piecewise plane fitting loss, we could enforce strong regularization on 3D structure. Although our slanted plane model is defined on disparity space, it has been proved that a plane in disparity space is still a plane in 3D space \cite{schneider2016semantically}. Mathematically, the disparity $d_p$ of each pixel $p$ is parameterized by a local plane,
\begin{equation}
d_p = a_pu+b_pv+c_p,
\end{equation}
where $(u,v)$ is the image coordinate, the triplet $(a_p,b_p,c_p)$ denotes the parameters of a local disparity plane.

Define $P$ as the matrix representation of pixel's homogeneous coordinates within a SLIC superpixel \cite{SLIC2012} with a dimension of $N\times3$ where $N$ is number of pixels within a segment, and denote $\mathbf{a}$ as the planar parameters. Given the current disparity predictions $\widehat{\mathbf{d}}$, we can estimate the plane parameter in closed-form via ${\bf a}^* = (P^TP)^{-1}P^T\widehat{\mathbf{d}}$. With the estimated plane parameter, the fitted planar disparities $\widetilde{\mathbf{d}} \in \mathbb{R}^N$ can be computed as $\widetilde{\mathbf{d}} = P\mathbf{a}^* = P(P^TP)^{-1}P^T\widehat{\mathbf{d}}$. 

Our plane fitting loss then can be defined as
\begin{equation}
\begin{split}
\mathcal{L}_{p} = \|\widehat{\mathbf{d}} - \widetilde{\mathbf{d}}\| = \|[I - P(P^TP)^{-1}P^T]\widehat{\mathbf{d}}\|.
\end{split}
\end{equation}

\begin{table*}[!ht]
\centering
\caption{\textbf{Quantitative results on the selected KITTI 141 subset.} We compare our LidarStereoNet with various state-of-the-art Lidar-Stereo fusion methods, where our proposed method outperforms all the competing methods with a wide margin.}
\begin{tabular}{l l c c c c c c c }
\toprule
Methods & Input & Supervised & Abs Rel & $>2$ px &  $>3$ px & $>5$ px　& $\delta<1.25$ & Density \\ \midrule
Input Lidar & Lidar & - &  - & 0.0572 & 0.0457 & 0.0375 & - & 7.27\%  \\ 
S2D \cite{Ma2018SparseToDense} & Lidar & Yes & 0.0665 & 0.0849 & 0.0659 & 0.0430 & 0.9626 & 100.00\% \\
SINet \cite{Uhrig2017THREEDV} & Lidar & Yes & 0.0659 & 0.0908 & 0.0660 & 0.0456 & 0.9576 & 100.00\% \\ \midrule
Probabilistic fusion \cite{Maddern2016}  & Stereo + Lidar & No & - & - & 0.0591 & - & - & 99.6\% \\
CNN Fusion \cite{Park2018} & Stereo + Lidar & Yes & - & - & 0.0484 & - & - & 99.8\% \\ \midrule
Our method & Stereo & No & {\bf0.0572}  & {\bf0.0540} & {\bf0.0345} & {\bf0.0220} & {\bf0.9731} & 100.00\% \\ 
Our method &Stereo + Lidar & No & {\bf 0.0350}  & {\bf 0.0287} & {\bf 0.0198} & {\bf 0.0126} &  {\bf 0.9872} & 100.00\% \\ 
\bottomrule
\end{tabular}
\label{tab:results_kitti141} 
\end{table*}

\begin{figure*}[!ht]
  \centering 
  \tabcolsep=0.06cm
  \begin{tabular}{c c c c} 
    \includegraphics[width=0.245\linewidth]{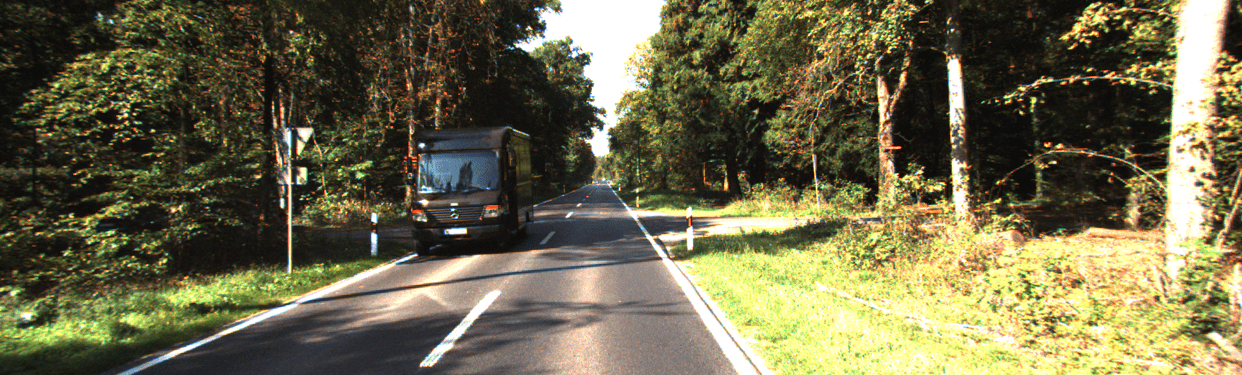} &
    \includegraphics[width=0.245\linewidth]{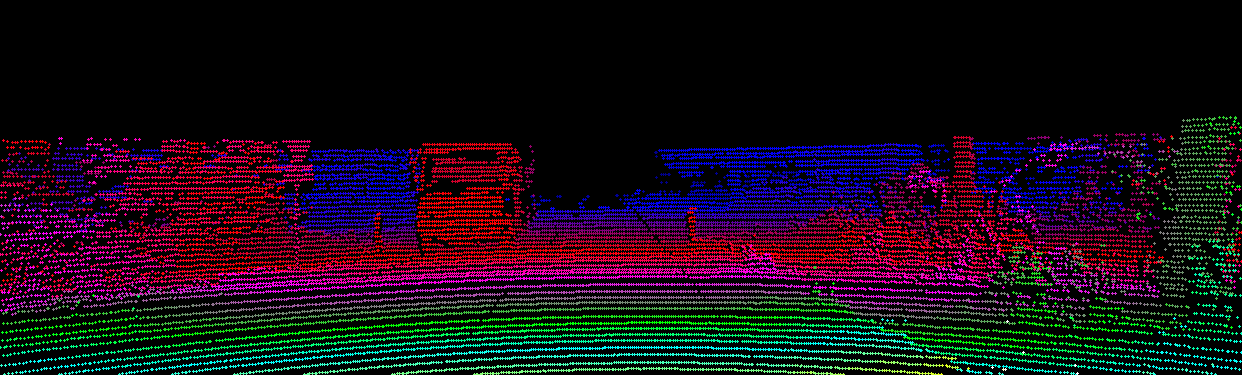}        &
    \includegraphics[width=0.245\linewidth]{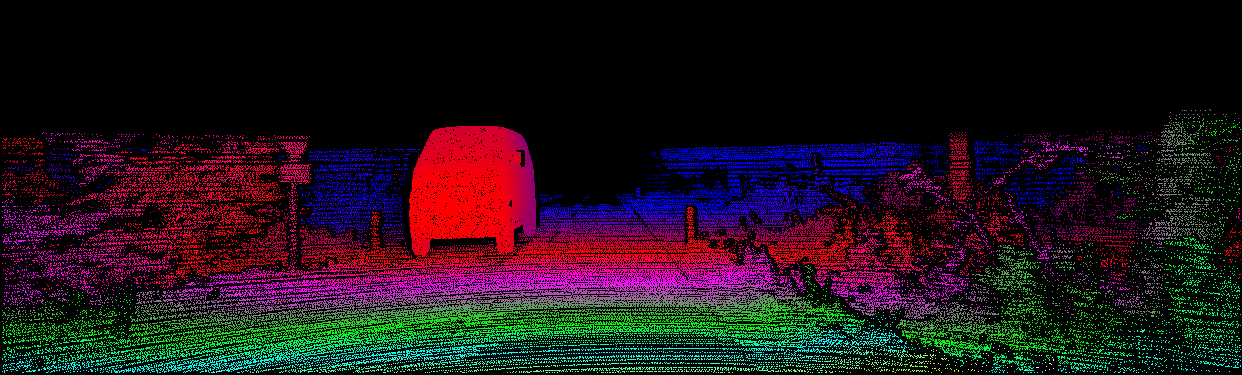}           &
    \includegraphics[width=0.245\linewidth]{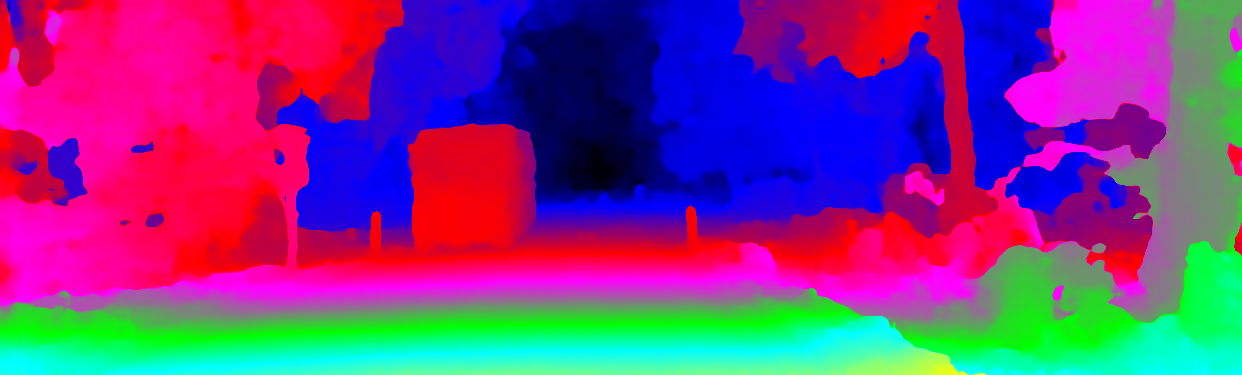} 
    \\
    (a) Input image & (b) Input lidar disparity & (c) Ground truth & (d) Ours
    \\
    \includegraphics[width=0.245\linewidth]{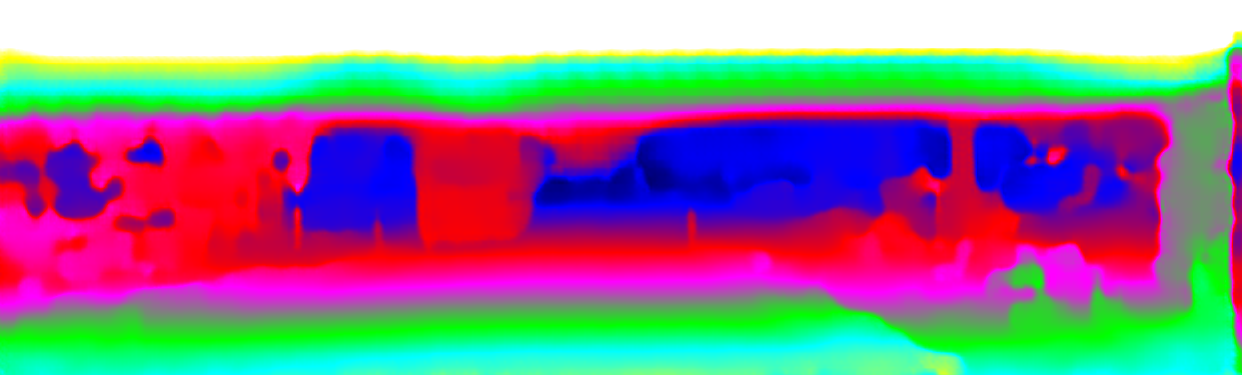}      &
    \includegraphics[width=0.245\linewidth]{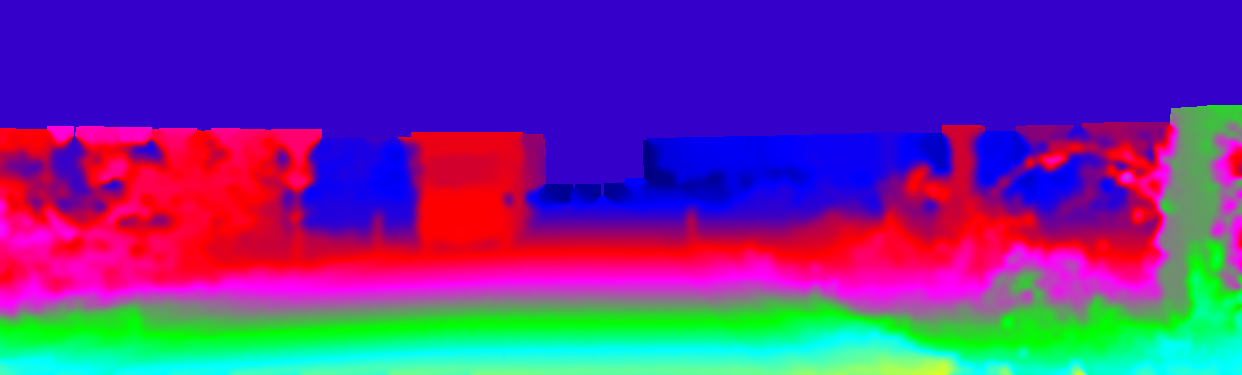}        &
    \includegraphics[width=0.245\linewidth]{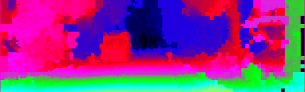} &
    \includegraphics[width=0.245\linewidth]{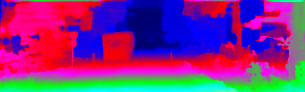}   
     \\
    (e)S2D \cite{Ma2018SparseToDense} & (f) SINet \cite{Uhrig2017THREEDV} & (g) Probabilistic fusion \cite{Maddern2016} & (h) CNN fusion \cite{Park2018} 
    \\    
  \end{tabular}  
  \caption{\textbf{Qualitative results of the methods from Tab.~\ref{tab:results_kitti141}.} Our method is trained on KITTI VO dataset and tested on the selected unseen KITTI 141 subset without any finetuning.} 
  \label{fig:results_kitti141_imgs} 
\end{figure*}

\section{Experiments}
We implemented our LidarStereoNet in Pytorch. All input images were randomly cropped to $256\times 512$ during training phases while we used their original size in inference. The typical processing time of our net was about 0.5 fps on Titan XP. We used the Adam optimizer with a constant learning rate of 0.001 and a batch size of 1. We performed a series of experiments to evaluate our LidarStereoNet on both real-world and synthetic datasets. In addition to analyzing the accuracy of depth prediction in comparison to previous work, we also conducted a series of ablation studies on different sensor fusing architectures and investigate how each component of the proposed losses contributes to the performance. 

\subsection{KITTI Dataset}

The KITTI dataset \cite{Geiger2012CVPR} is created to set a benchmark for autonomous driving visual systems. It captures depth information from a Velodyne HDL-64E Lidar and corresponding stereo images from a moving platform. 
They use a highly accurate inertial measurement unit to accumulate 20 frames of raw Lidar depth data in a reference frame and serves as ground truth for the stereo matching benchmark. 
In KITTI 2015 \cite{Menze2015ISA}, they also take moving objects into consideration. The dynamic objects are first removed and then re-inserted by fitting CAD models to the point clouds, resulting in a clean and dense ground truth for depth evaluation.

\vspace{-3mm}
\paragraph{Dataset Preparation}
After these processes, the raw Lidar points and the ground truth differ significantly in terms of outliers and density as shown in Fig.~\ref{fig:lidar_error}. In raw data, due to the large displacement between the Lidar and the stereo cameras \cite{schneider2016semantically}, boundaries of objects may not perfectly align when projecting Lidar points onto image planes. Also, since Lidar system scans depth in a line by line order, it will create a rolling shutter effect on the reference image, especially for a moving platform. Instead of heuristically removing measurements, our method is able to omit these outliers automatically which is evidently shown in Fig.~\ref{fig:lidar_error} and Fig.~\ref{fig:lidar_error2}. 

We used the KITTI VO dataset \cite{Geiger2012CVPR} as our training set. We sorted all 22 KITTI VO sequences and found 7 frames from sequence 17 and 20 having corresponding frames in the KITTI 2015 training set. Therefore we excluded these two sequences and used the remaining 20 stereo sequences as our training dataset. Our training dataset contains 42104 images with a typical image resolution of $1241\times376$. To obtain sparse disparities inputs, we projected raw Lidar points onto left and right images using provided extrinsic and intrinsic parameters and converted the raw Lidar depths to disparities. Maddern \etal \cite{Maddern2016} also traced 141 frames from KITTI raw dataset that have corresponding frames in the KITTI 2015 dataset and reported their results on this subset. For consistency, we used the same subset to evaluate our performance and utilize the 6 frames from KITTI VO dataset as our validation set (we excluded 1 frame that overlaps the KITTI 141 subset from our validation).

\vspace{-3mm}
\paragraph{Comparisons with State-of-the-Art}
We compared our results with depth completion methods and Lidar-stereo fusion methods using depth metrics from \cite{Eigen2014} and bad pixel ratio disparity error from KITTI \cite{Menze2015ISA}. We also provide a comparison of our method and stereo matching methods in the supplemental material. 

For depth completion, we compared with S2D \cite{Ma2018SparseToDense} and SINet \cite{Uhrig2017THREEDV}. In our implementation of S2D and SINet, we trained them on KITTI depth completion dataset \cite{Uhrig2017THREEDV}. From 151 training sequences, we excluded 28 sequences that overlaps with KITTI 141 dataset and used the remaining 123 sequences to train these networks from scratch in a supervised manner. As a reference, we computed the error rate of the input Lidar. It is worth noting that our method increases the disparity density from less than 7.3\% to 100\% while reducing the error rate by a half. 

We also compared our method with two existing Lidar-Stereo fusion methods: Probabilistic fusion \cite{Maddern2016} and CNN fusion \cite{Park2018} and outperforms them with a large margin. Quantitative comparison between our method and the competing state-of-the-art methods is reported in Tab.~\ref{tab:results_kitti141}. We can clearly see that our self-supervised LidarStereoNet achieves the best performance throughout all the metrics evaluated. Note that, our method even outperforms recent supervised CNN based fusion method \cite{Park2018} with a large margin. More qualitative evaluations of our method in challenging scenes are provided in Fig.~\ref{fig:results_kitti141_imgs}.
These results demonstrate the superiority of our method that can effectively leverage the complementary information between Lidar and stereo images.

\begin{figure}[t]
    \centering
    \includegraphics[width=1\linewidth,height=0.50\linewidth]{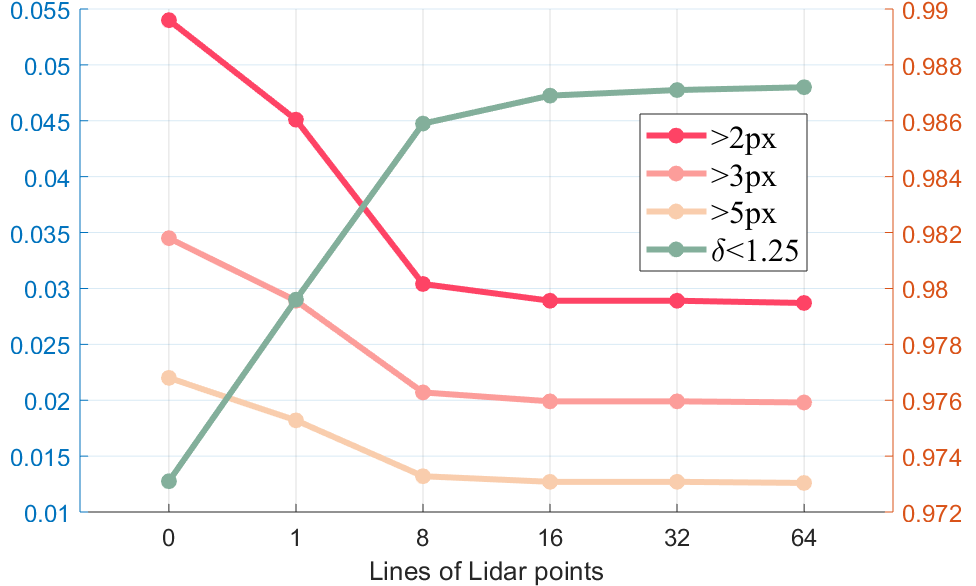}
\caption{\textbf{Test results of our network on the selected KITTI 141 subset with varying levels of input Lidar points sparsity.} Left column: lower is better; right column: higher is better.} 
\label{fig:sparse_lidar_results }
\end{figure}

\vspace{-3mm}
\paragraph{On Input Sparsity}
Thanks to the nature of deep network and sparsity invariant convolution, our LidarStereoNet can handle Lidar input of varying density, ranging from no Lidar input to 64 lines input. To see this trend, we downsampled the vertical and horizontal resolution of the Lidar points. As shown in Fig.~\ref{fig:sparse_lidar_results }, our method performs equally well when using 8 or more lines of Lidar points. Note that even when there are no Lidar points as input (in this case, the problem becomes a pure stereo matching problem), our method still outperforms SOTA stereo matching methods. 

\begin{table}[!ht]
\centering
\caption{\textbf{Ablation study on the feedback loop} Type 1 and Type 2 show the performance only use the Core Architecture without and with removing error Lidar points from the input, while Full model means our proposed feedback loop.}
\tabcolsep=0.28cm
\begin{tabularx}{\columnwidth}{c c c c c c}
\toprule
Methods & Abs Rel & $>2$ px &  $>3$ px & $>5$ px \\ 
\midrule
Type 1 & 0.0539 & 0.0411 & 0.0310 & 0.0229 \\ 
Type 2 & 0.0468 & 0.0401 & 0.0302 & 0.0226 \\ 
Full model & {\bf 0.0350} & {\bf 0.0287} & {\bf 0.0198} & {\bf 0.0126} \\ 
\bottomrule
\end{tabularx}
\label{tab:kitti_feedback}
\end{table}

\subsection{Ablation Study}
In this section, we perform ablation studies to evaluate the importance of our feedback loop and proposed losses. Notably, all ablation studies on losses and fusion strategies are evaluated on Core Architecture only in order to reduce the randomness introduced by our feedback loop module.

\vspace{-3mm}
\paragraph{Importance of the feedback loop}
We evaluate the importance of the feedback loop in two aspects. One is to remove the error points from the back-end, \ie the loss computation part. The other is to remove them from the input. In our baseline model (\emph{Type 1}), we use raw Lidar as our input and compute the Lidar loss on them. For \emph{Type 2} model, we also use the raw Lidar as input but compute the Lidar loss only on cleaned Lidar points. Our full model uses clean Lidar points in both parts. As shown in Tab.~\ref{tab:kitti_feedback}, removing errors in the back-end can improve the performance by $2.58\%$. However, using cleaned Lidar points as input can boost the performance in $34.44\%$ in $>3 px$ metric, which demonstrates the importance of our feedback loop module. 


\begin{figure}
    \centering
    \includegraphics[width=0.495\linewidth]{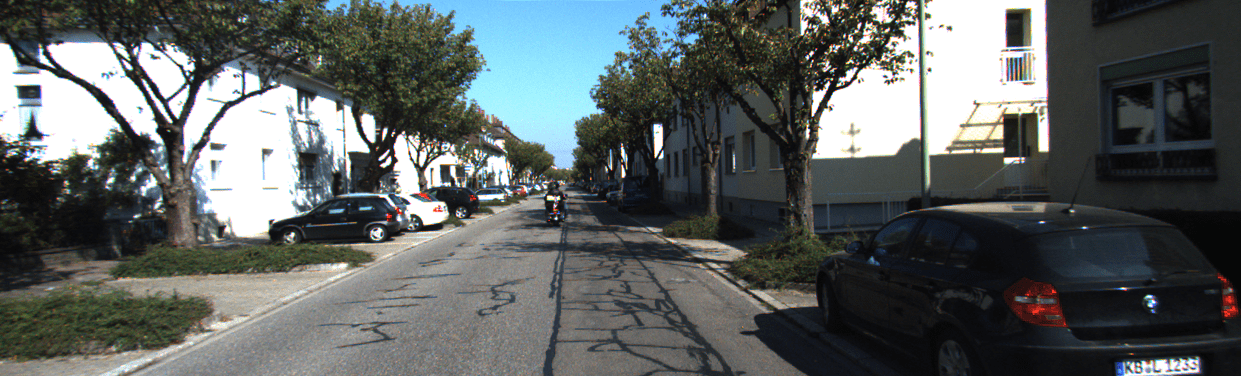}
    \includegraphics[width=0.495\linewidth]{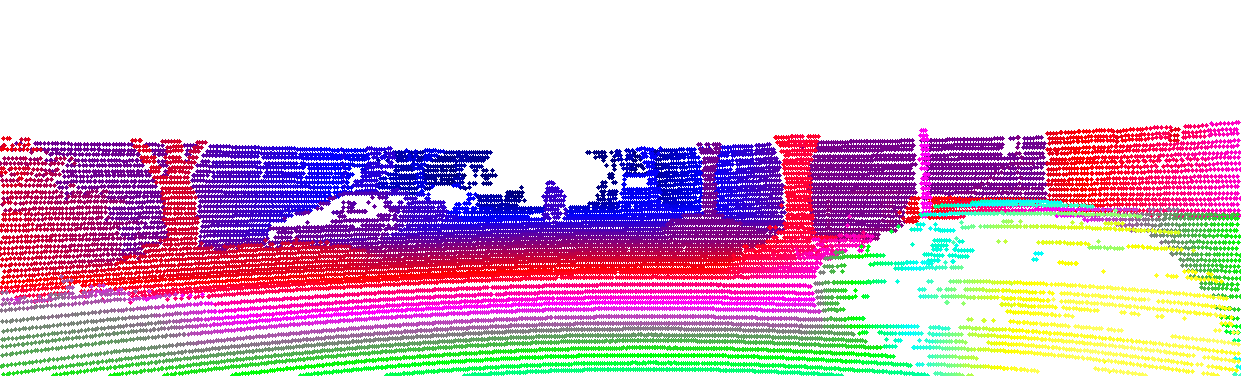} 
    \includegraphics[width=0.495\linewidth]{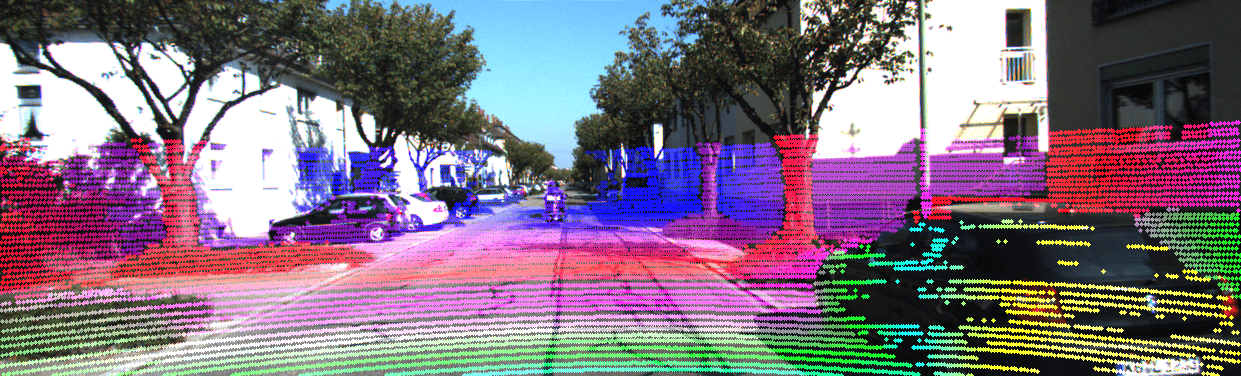}
    \includegraphics[width=0.495\linewidth]{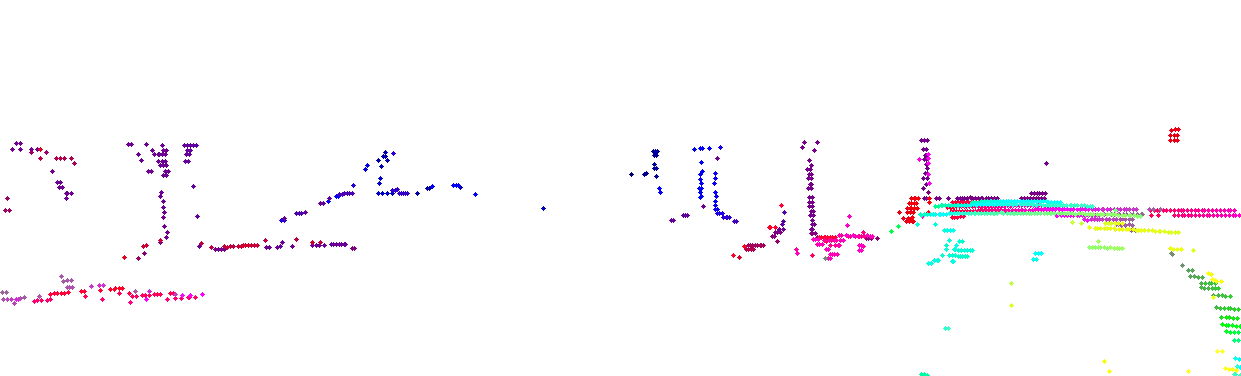} 
    \includegraphics[width=0.495\linewidth]{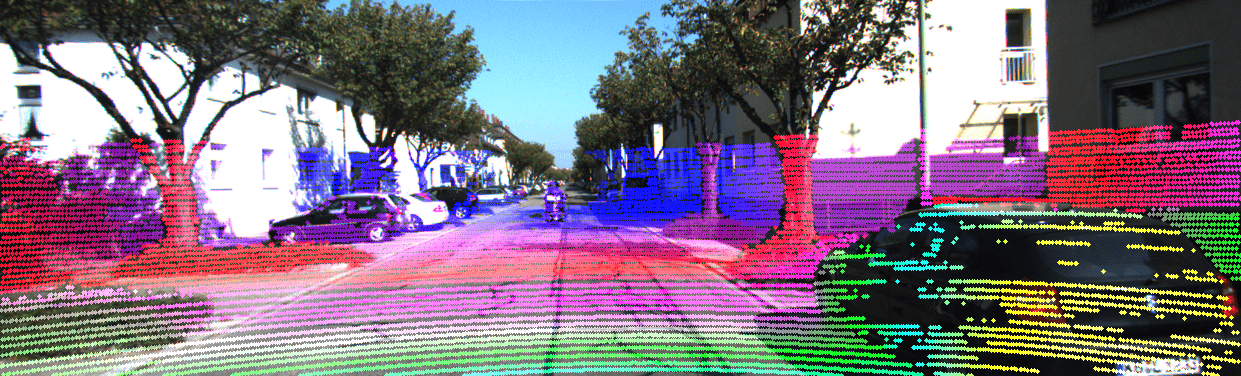}
    \includegraphics[width=0.495\linewidth]{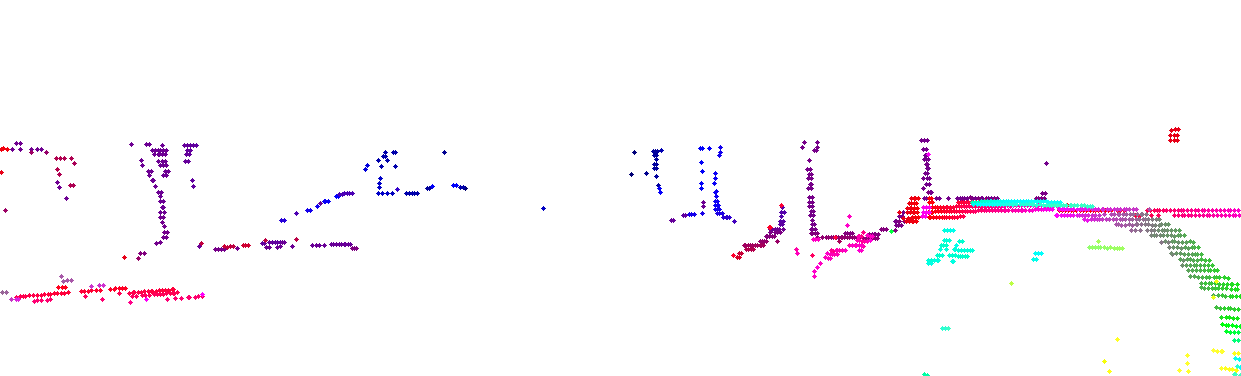}
    \caption{\textbf{Gradually cleaned input Lidar points.} From top to bottom, left column: left image, cleaned Lidar points at the $2^{nd}$ epoch, cleaned Lidar points at the $5^{th}$ epoch; Right column: raw Lidar points, error points find at the $2^{nd}$ epoch, error points find at the $5^{nd}$ epoch. Note that the error measurements on the right car have been gradually removed.}
    \label{fig:clean_lidar}
\end{figure}

\vspace{-3mm}
\paragraph{Comparing different loss functions}
Tab.~\ref{tab:kitti_loss_comparison} shows the performance gain with different losses. As we can see, when only using Lidar points as supervision, its performance is affected by the outliers in Lidar measurements. Adding a warping loss can reduce the error rate from $4.71\%$ to $3.02\%$. Adding our proposed plane fitting loss can further reduce the metric from $3.02\%$ to $2.73\%$. In the supplementary material, we further compare our soft slanted plane model and a hard plane fitting model. The soft one achieves better performance.

\begin{table}[!ht]
\centering
\caption{\label{tab:kitti_loss_comparison} {\textbf{Evaluation of different loss functions.} $\mathcal{L}_{w}$, $\mathcal{L}_{s}$, $\mathcal{L}_{l}$ and $\mathcal{L}_{p}$ represent warping loss, smoothness loss, Lidar loss and plane fitting loss separately.}}
\tabcolsep=0.15cm
\begin{tabularx}{\columnwidth}{l c c c c c }
\toprule
Loss & Abs Rel & $>2$ px &  $>3$ px &  $>5$ px \\ \midrule
$\mathcal{L}_{l}$ & 0.0555 & 0.0733 & 0.0471 &  0.0296  \\ $\mathcal{L}_{w}$ + $\mathcal{L}_{s}$ & 0.0628 & 0.0940 & 0.0637 & 0.0405 \\ 
$\mathcal{L}_{w}$ + $\mathcal{L}_{s}$ + $\mathcal{L}_{l}$ & 0.0565 & 0.0401 & 0.0302 & 0.0226\\ 
$\mathcal{L}_{w}$ + $\mathcal{L}_{s}$ + $\mathcal{L}_{l}$ + $\mathcal{L}_{p}$ & {\bf 0.0468}  & {\bf 0.0393} & {\bf 0.0276} & {\bf 0.0201}  \\ 
\bottomrule
\end{tabularx}
\end{table}

\vspace{-3mm}
\paragraph{Comparing different fusion strategies}
Considering the problem of utilizing sparse depth information, one no-fusion approach will be directly using Lidar measurements for supervisions. As shown in Tab.~\ref{tab:kitti_fusion_strategy}, its performance is affected by the misaligned Lidar points and it has a relatively high error rate of $4.10\%$. The second method is to leverage the depth as a fourth channel additionally to the RGB images. We term it an early fusion strategy. As shown in Tab.~\ref{tab:kitti_fusion_strategy}, it has the worst performance among the baselines. This may be due to the incompatible characteristics between RGB images and depth maps thus the network is unable to handle well within a common convolution layer. Our late fusion strategy achieves the best performance among them.

\begin{table}
\centering
\caption{\label{tab:kitti_fusion_strategy} {\textbf{Comparison of different fusion strategies.} }}
\tabcolsep=0.2cm
\begin{tabularx}{\columnwidth}{ c c c c c c }
\toprule
Methods & Abs Rel & $>2$ px &  $>3$ px & $>5$ px \\ \midrule
No Fusion & 0.0555 & 0.0733 & 0.0471 & 0.0296 \\ 
Early fusion & 0.0644& 0.0667 & 0.0526 & 0.0398\\ 
Our method & {\bf 0.0468} & {\bf 0.0393} & {\bf 0.0276} & {\bf 0.0201}\\ 
\bottomrule
\end{tabularx}
\end{table}


\subsection{Generalizing to Other Datasets}
To illustrate that our method can generalize to other datasets, we compare our method to several methods on the Synthia dataset \cite{RosCVPR16}. Synthia contains 5 sequences under different scenarios. And for each scenario, they capture images under different lighting and weather conditions such as Spring, Winter, Soft-rain, Fog and Night. We show quantitative results of experiments in Tab.~\ref{tab:synthia} and qualitative results are provided in the supplementary material.

\begin{figure}[t]
    \centering
    \includegraphics[width=1\linewidth,height=0.50\linewidth]{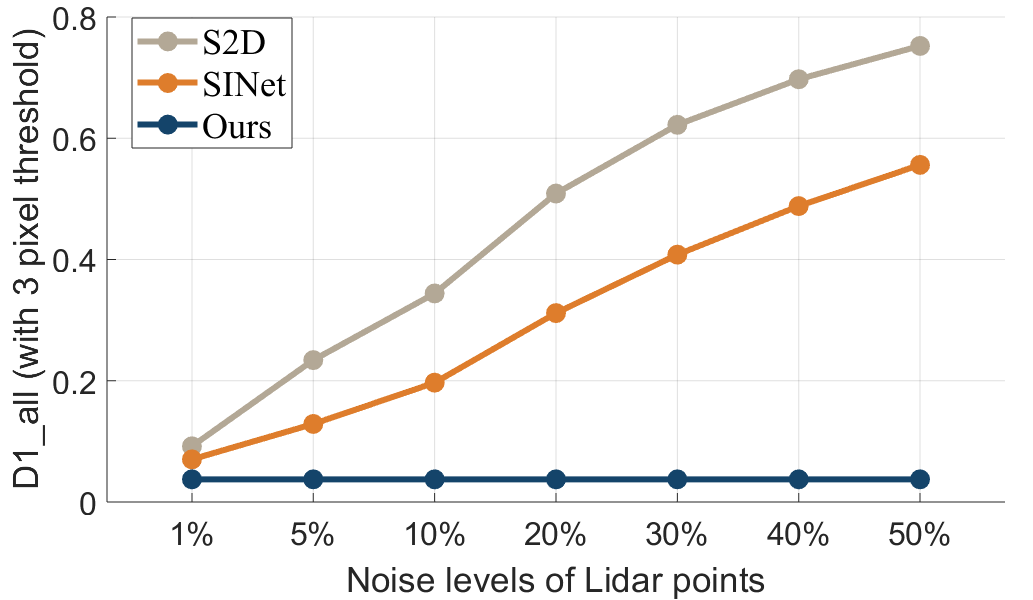}
\caption{\textbf{Ablation study on noise resistance on Synthia dataset.} Our method has a consistent performance while the others have a notable performance drop.} 
\label{fig:syn_noise_lidar }
\end{figure}

For sparse disparity inputs, we randomly selected 10\% of full image resolution. As discussed before, projected Lidar points have misalignment with stereo images in KITTI dataset. To simulate the similar interference, we add various density levels of Gaussian noise to sparse disparity maps. As shown in Fig.~\ref{fig:syn_noise_lidar }, our proposed LidarStereoNet adapts well to the noisy input disparity maps, while S2D \cite{Ma2018SparseToDense} fails to recover disparity information. 

\begin{table}[h]
\centering
\caption{\label{tab:synthia} {\textbf{Quantitative results on the Synthia dataset.}}}
\tabcolsep=0.26cm
\begin{tabularx}{\columnwidth}{ c c c c c c }
\toprule
Methods & Abs Rel & $>2$ px &  $>3$ px & $>5$ px \\ \midrule
SPS-ST \cite{Yamaguchi14}  &  0.0475 &  0.0980 &  0.0879 &  0.0713 \\ 
S2D \cite{Ma2018SparseToDense} & 0.0864 & 0.5287 & 0.4414 & 0.270 \\ 
SINet \cite{Uhrig2017THREEDV} & {\bf 0.0290} & 0.0642 & 0.0472 & {\bf0.0283} \\ 
Our method & 0.0334 & {\bf 0.0446} & {\bf 0.0373} &  0.0299\\
\bottomrule
\end{tabularx}
\end{table}

\section{Conclusion}
In this paper, we have proposed an unsupervised end-to-end learning based Lidar-Stereo fusion network ``LidarStereoNet'' for accurate 3D perception in real world scenarios. To effectively handle noisy Lidar points and misalignment between sensors, we presented a novel ``Feedback Loop'' to sort out clean measurements by comparing output stereo disparities and input Lidar points. We have also introduced a piecewise slanted plane fitting loss to enforce strong 3D structural regularization on generated disparity maps. Our LidarStereoNet does not need ground truth disparity maps for training and has good generalization capabilities. Extensive experiments demonstrate the superiority of our approach, which outperforms state-of-the-art stereo matching and depth completion methods with a large margin. Our approach can reliably work even when Lidar points are completely missing. In the future, we plan to extend our method to other depth perception and sensor fusion scenarios.


\noindent
\textbf{Acknowledgement}
Y. Dai (daiyuchao@gmail.com) is the corresponding author. This research was supported in part by Australia Centre for Robotic Vision, Data61 CSIRO, the Natural Science Foundation of China grants (61871325, 61420106007) the Australian Research Council (ARC) grants (LE190100080, CE140100016, DP190102261, DE140100180). The authors are grateful to the GPUs donated by NVIDIA. 
{\small
\bibliographystyle{ieee}
\bibliography{LidarStereo-Reference}
}

\clearpage
\onecolumn
\begin{center}
\LARGE
\bf
    Noise-Aware Unsupervised Deep Lidar-Stereo Fusion\\--Supplementary Material -- 
\end{center}

\begin{abstract}
In this supplementary material, we provide our detailed network structure, qualitative comparison of hard and soft slanted plane constraint, qualitative and quantitative comparison to stereo matching algorithms and qualitative results on the Synthia dataset. 
\end{abstract}
\setcounter{section}{0}
\section{Detailed Network Structure}  
The core architecture of our LidarStereoNet contains three blocks: 1) Feature extraction and fusion; 2) Feature matching, and 3) Disparity computing. We provide the detailed structure of the feature extraction and fusion block in Table \ref{tab:featureext}. The feature matching block and disparity computing block share the same structures with PSMnet \cite{chang2018pyramid}.

\begin{table}[!htp]
\small
\caption{\label{tab:featureext} \small Feature extraction and fusion block architecture, where \textbf{k}, \textbf{s}, \textbf{chns} represent the kernel size, stride and the number of the input and the output channels. We use ``+'' to represent feature concatenation.}
\tabcolsep=0.5cm
\centering
\begin{tabular}{|c|l|c|c|}
\hline

\multicolumn{4}{|c|}{\textbf{Lidar feature extraction}}                                                      \\ \hline
\textbf{layer} & \textbf{k} , \textbf{s} & \textbf{chns}  & \textbf{input} \\ \hline
conv\_s1          & 11$\times$11, 1      & 1/16           & disparity        \\
conv\_s2          & 7$\times$7, 2          & 16/16           & conv\_s1         \\
conv\_s3          & 5$\times$5, 1          & 16/16           & conv\_s2         \\
conv\_s4          & 3$\times$3, 2          & 16/16           & conv\_s3         \\
conv\_s5          & 3$\times$3, 1          & 16/16           & conv\_s4         \\
conv\_mask        & 1$\times$1, 1          & 17/16           & conv\_s5+mask    \\ \hline

\multicolumn{4}{|c|}{\textbf{Stereo feature extraction}}                                                      \\ \hline
\textbf{layer} & \textbf{k} , \textbf{s} & \textbf{chns}  & \textbf{input} \\ \hline
conv0\_1          & 3$\times$3, 2          & 3/32        & image           \\ 
conv0\_2          & 3$\times$3, 1          & 32/32       & conv0\_1        \\ 
conv0\_3          & 3$\times$3, 1          & 32/32       & conv0\_2       \\ 

conv1\_n    &  \bigg[ \begin{tabular}{l}
3$\times$3, 1  \\
3$\times$3, 1
\end{tabular}  \bigg]$\times$3           & 32/32     & conv0\_3          \\ 

conv2\_1    &  \bigg[ \begin{tabular}{l}
3$\times$3, 2  \\
3$\times$3, 1  
\end{tabular}  \bigg]             & \bigg[\begin{tabular}{c} 32/64 \\ 64/64 \end{tabular} \bigg]    & conv1\_3          \\ 
conv2\_n    &  \bigg[ \begin{tabular}{l}
3$\times$3, 1  \\
3$\times$3, 1
\end{tabular}  \bigg]$\times$15           & 64/64     & conv2\_1          \\ 

conv3\_1    &  \bigg[ \begin{tabular}{l}
3$\times$3, 1  \\
3$\times$3, 1
\end{tabular}  \bigg]            & \bigg[\begin{tabular}{c} 64/128 \\ 128/128 \end{tabular} \bigg]     & conv2\_16          \\ 
conv3\_n    &  \bigg[ \begin{tabular}{l}
3$\times$3, 1  \\
3$\times$3, 1
\end{tabular}  \bigg]$\times$2           & 128/128     & conv3\_1          \\ 

conv4\_n    &  \bigg[ \begin{tabular}{l}
3$\times$3, 1  \\
3$\times$3, 1
\end{tabular}  \bigg]$\times$3           & 128/128     & conv3\_3          \\ 

branch1       & 64$\times$64, 64          & 128/32       & conv4\_3       \\ 
branch2       & 32$\times$32, 32          & 128/32       & conv4\_3       \\ 
branch3       & 64$\times$16, 16          & 128/32       & conv4\_3       \\ 
branch4       & 8$\times$8, 8            & 128/32       & conv4\_3       \\ 
lastconv      & \bigg[ \begin{tabular}{c}
3$\times$3, 1  \\
1$\times$1, 1
\end{tabular}  \bigg]   & \bigg[ \begin{tabular}{c} 320/128 \\128/32 \end{tabular} \bigg]  &\begin{tabular}{c} conv2\_16+conv4\_3 \\+branch1+branch2 \\+branch3+branch4 \end{tabular}    \\ 

\hline
\multicolumn{4}{|c|}{\textbf{Feature fusion}}           \\ \hline
\multicolumn{4}{|c|}{lastconv + conv\_mask}  \\

\hline         
\end{tabular}
\end{table}

\section{Hard versus Soft Plane Fitting}
There are two kinds of plane fitting constraints. Conventional CRF based methods use one slanted plane model to describe all disparities in one segment, \ie, disparities insides one segment exactly obeys one slanted plane model. We term it as ``Hard'' plane fitting constraint. Our method, on the other hand, only applies this term as part of the whole optimization target. In other words, we only require the recovered disparities to fit a plane in a segment if possible but it can still be balanced by other loss terms. 

\begin{figure*}[!ht]
    \centering
    \tabcolsep=0.1cm
    \begin{tabular}{c c}
    \includegraphics[width=0.48\linewidth]{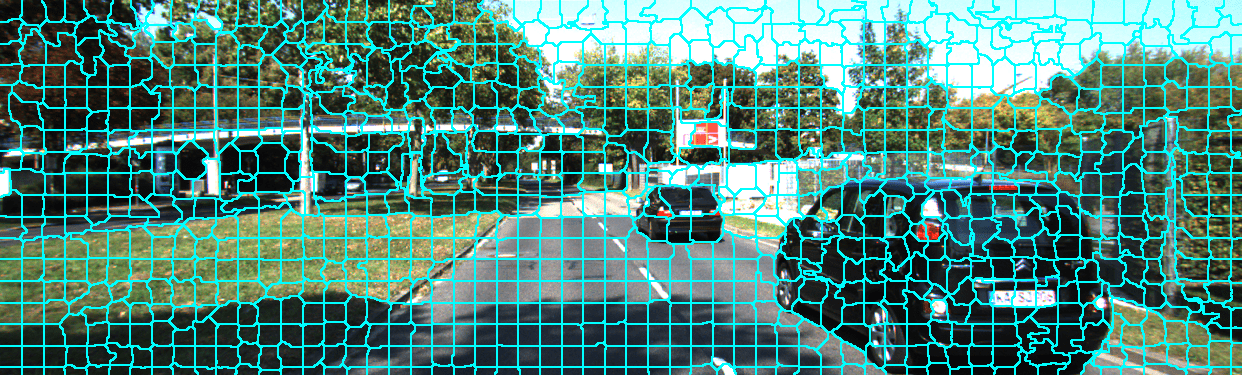}  &
    \includegraphics[width=0.48\linewidth]{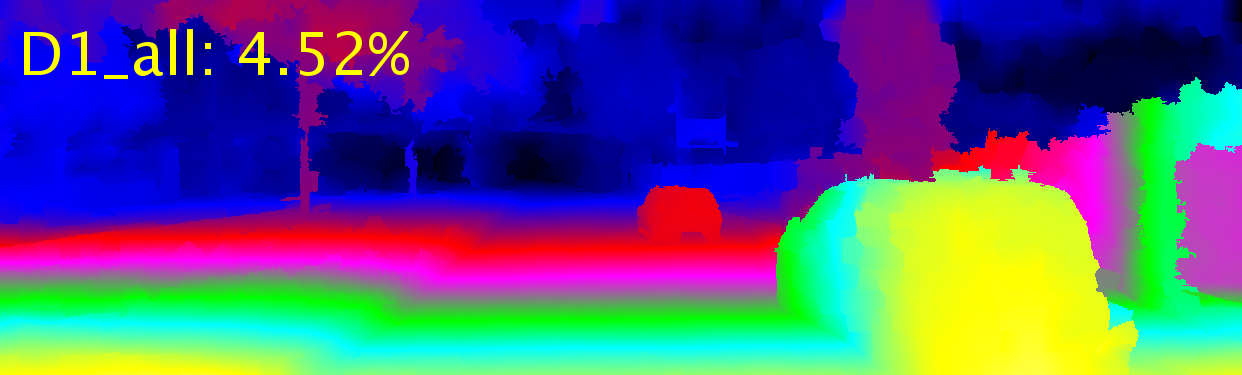} \\
    SLIC segments & Hard constraint result 
    \\
    \includegraphics[width=0.48\linewidth]{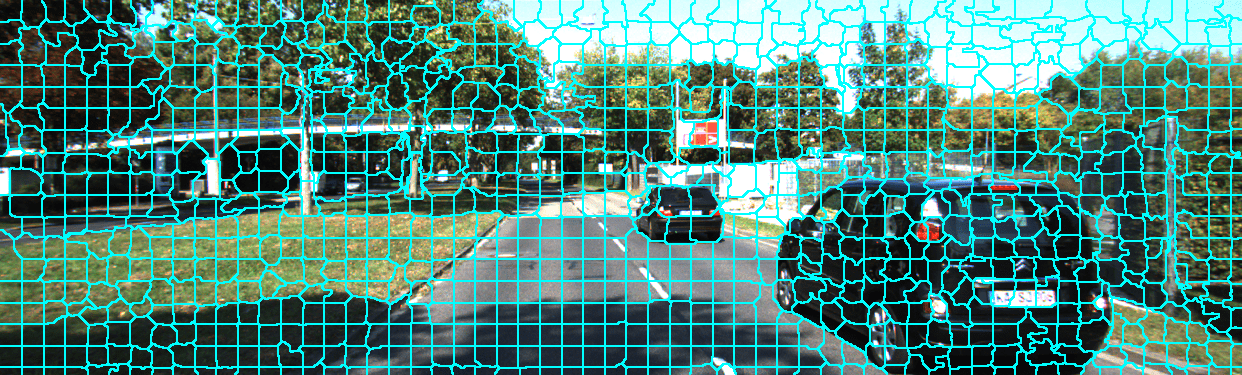}      &
    \includegraphics[width=0.48\linewidth]{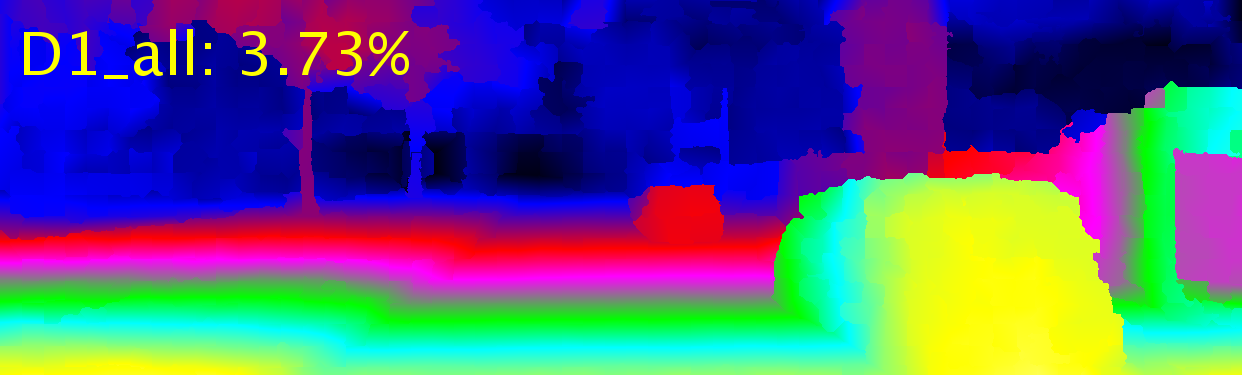} \\
    Stereo SLIC segments &  Hard constraint result
    \\
    \includegraphics[width=0.48\linewidth]{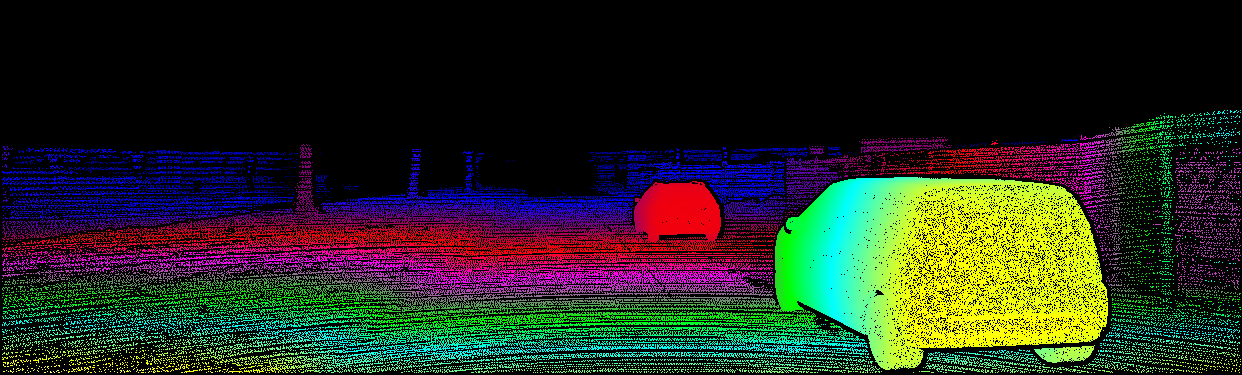}      &
    \includegraphics[width=0.48\linewidth]{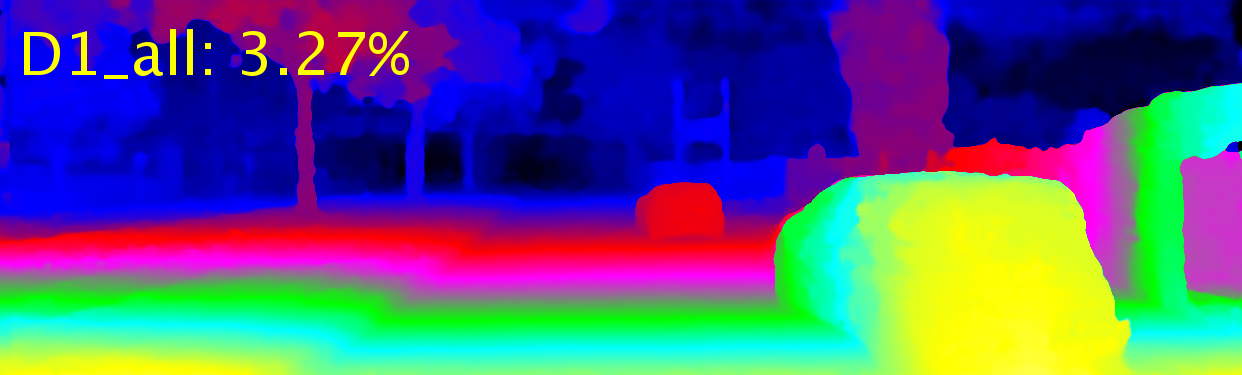} \\
    Ground truth disparity & Soft constraint result 
    \\
    \end{tabular}
    \caption{\textbf{Comparison of soft and hard constraints on slanted plane model with different superpixel segmentation methods.}  Note that our recovered disparity map has more aligned boundaries with the color image.}
    \label{fig:hard_soft}
\end{figure*}

Fig.~\ref{fig:hard_soft} illustrates the difference between our soft constraint and the CRF-style hard constraint in a recovered disparity map. As can be seen in Fig.~\ref{fig:hard_soft}, strictly applying the slanted plane model in recovered disparity map decreases its performance from $3.27\%$ to $3.97\%$ and it is very sensitive to segments as well. By switching segments from Stereo SLIC to SLIC, its performance further decreases from $3.97\%$ to $4.52\%$.

\begin{figure*}[!ht]
    \centering
    \tabcolsep=0.03cm
    \begin{tabular}{c c c c c c}
    \includegraphics[width=0.195\linewidth]{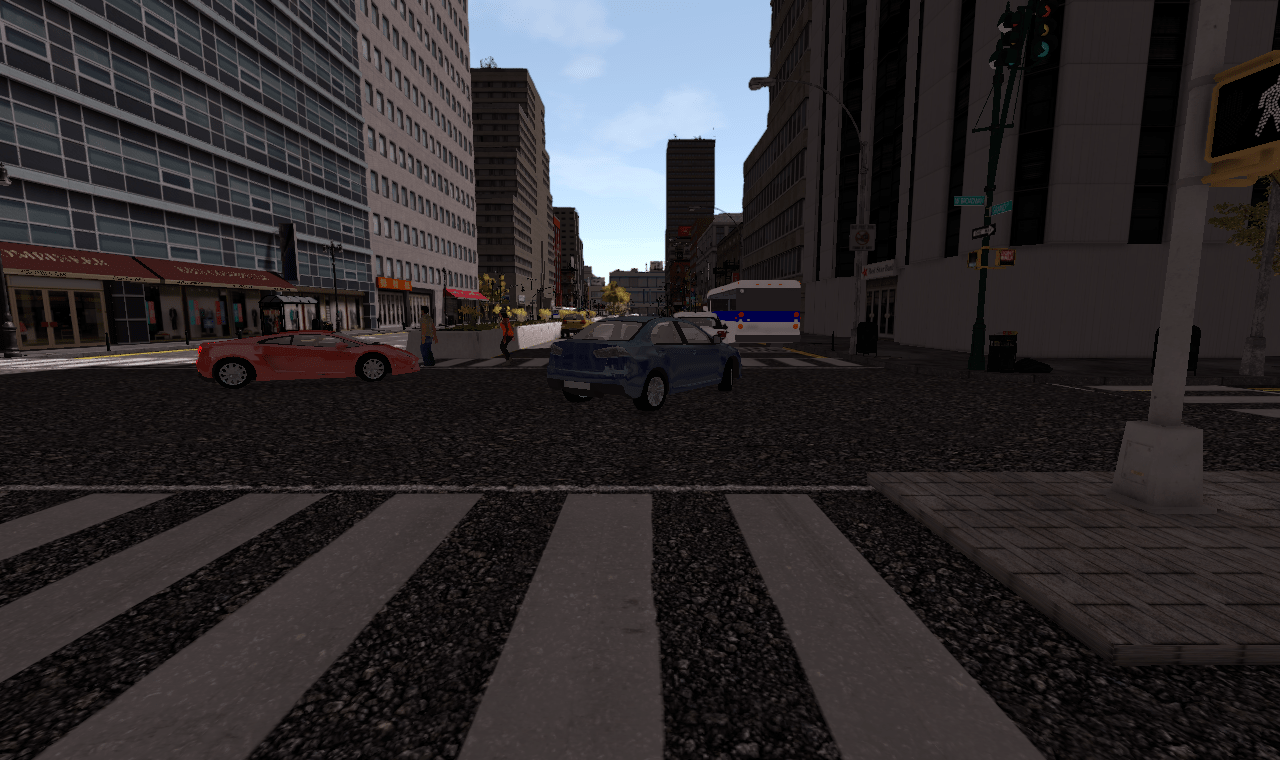}          &
    \includegraphics[width=0.195\linewidth]{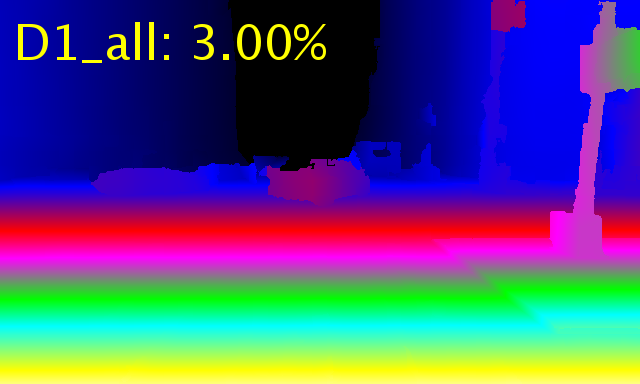} &
    \includegraphics[width=0.195\linewidth]{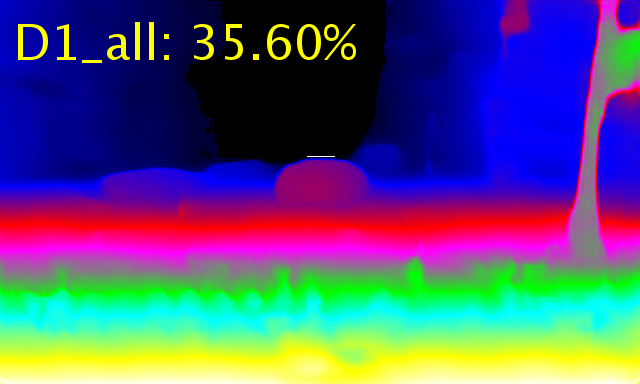}  &
    \includegraphics[width=0.195\linewidth]{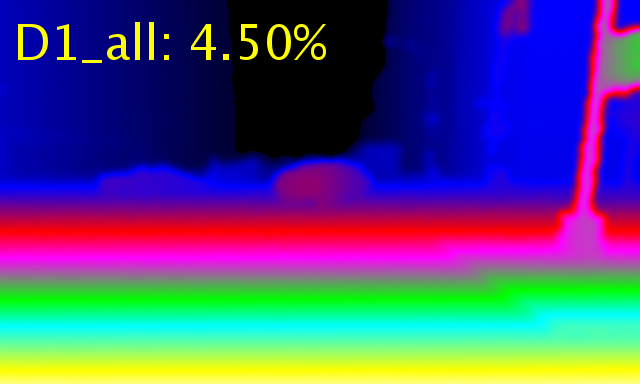} &
    \includegraphics[width=0.195\linewidth]{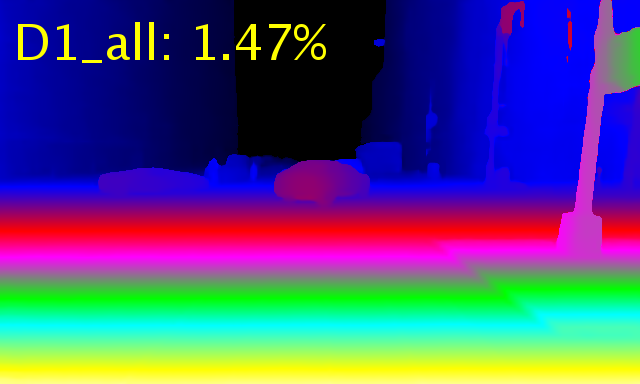} \\
    \includegraphics[width=0.195\linewidth]{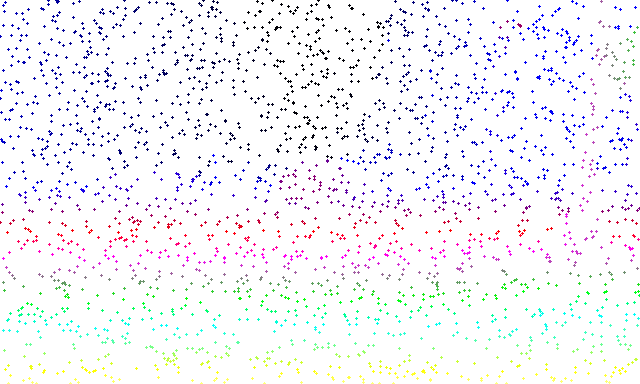} &
    \includegraphics[width=0.195\linewidth]{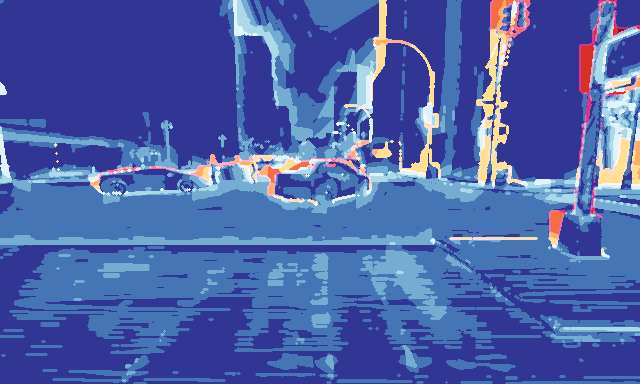}  &
    \includegraphics[width=0.195\linewidth]{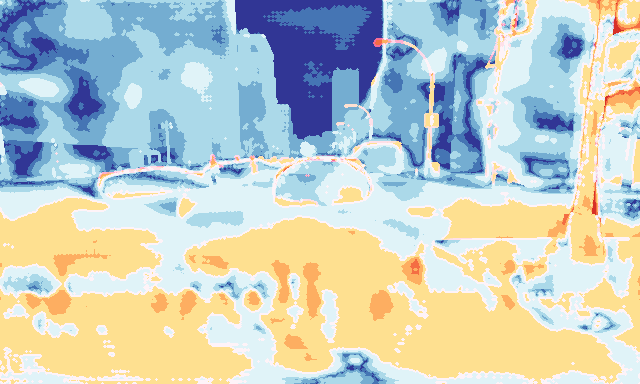}   &
    \includegraphics[width=0.195\linewidth]{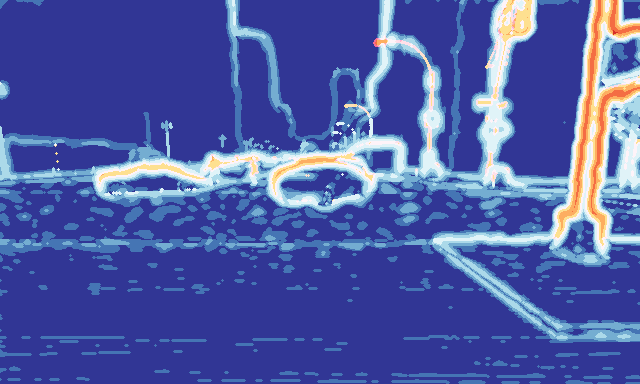}  &
    \includegraphics[width=0.195\linewidth]{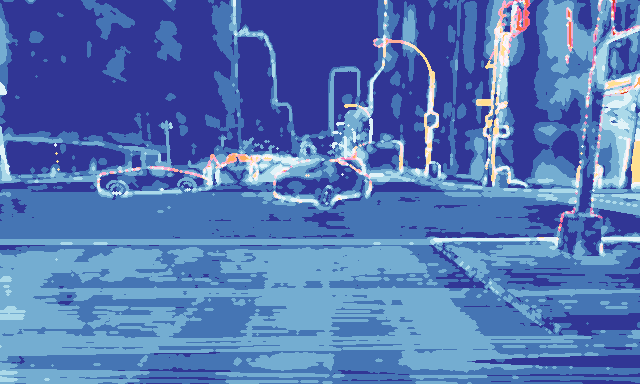}     \\
    Sparse disparity & SPS-ST \cite{Yamaguchi14} & S2D \cite{Ma2018SparseToDense} & SINet \cite{Uhrig2017THREEDV} & Ours
    \end{tabular}
    \caption{\textbf{Qualitative results on the Synthia dataset.} The first raw is the colorized disparity results, and the second row is the corresponding error maps.}
    \label{fig:synthia_imgs}
\end{figure*}


\section{Comparisons with STOA stereo matching methods}
For the sake of completeness, we provide qualitative and quantitative comparisons with state-of-the-art stereo matching methods. We choose SPS-ST \cite{Yamaguchi14}, MC-CNN\cite{Zbontar2016}, PSMnet \cite{chang2018pyramid} and SsSMnet\cite{zhong2017self} for reference. Note that the SPS-ST method is a traditional (non-deep) method, and its meta-parameters were tuned on KITTI dataset. For deep MC-CNN we used a model which was firstly trained on Middlebury dataset and for PSMnet we used the model that was trained on SceneFlow \cite{Mayer2016CVPR} dataset and the model (``-ft'') that we fine-tuned on KITTI VO dataset. We also compared our method with state-of-the-art self-supervised stereo matching network SsSMnet \cite{zhong2017self}.

\begin{table*}[!ht]
\centering
\caption{\textbf{Quantitative comparison on the selected KITTI 141 subset.} We compare our LidarStereoNet with various state-of-the-art stereo matching methods, where our proposed method outperforms all the competing methods with a wide margin.}
\begin{tabular}{l l c c c c c c c }
\toprule
Methods & Input & Supervised & Abs Rel & $>2$ px &  $>3$ px & $>5$ px　& $\delta<1.25$ & Density \\ \midrule
MC-CNN \cite{Zbontar2016} & Stereo & Yes & 0.0798 & 0.1070 & 0.0809 & 0.0555 & 0.9472 & 100.00\%\\
PSMnet \cite{chang2018pyramid} & Stereo & Yes & 0.0807 & 0.2480 & 0.1460 & 0.0639 & 0.9399 & 100.00\%\\
PSMnet-ft \cite{chang2018pyramid}  & Stereo & Yes & 0.0609 & 0.0635 & 0.0410 & 0.0277 & 0.9689& 100.00\%\\
SPS-ST \cite{Yamaguchi14} & Stereo & No & 0.0633 & 0.0702 & 0.0413 & 0.0265 & 0.9660 & 100.00\% \\ 
SsSMnet \cite{zhong2017self} & Stereo & No & 0.0619 &  0.0743 & 0.0498 & 0.0334 & 0.9633 & 100.00\% \\ \midrule
Our method & Stereo & No & {\bf0.0572}  & {\bf0.0540} & {\bf0.0345} & {\bf0.0220} & {\bf0.9731} & 100.00\% \\ 
Our method &Stereo + Lidar & No & {\bf 0.0350}  & {\bf 0.0287} & {\bf 0.0198} & {\bf 0.0126} &  {\bf 0.9872} & 100.00\% \\ 
\bottomrule
\end{tabular}
\label{tab:results_kitti141_add} 
\end{table*}

\begin{figure*}[!ht]
  \centering 
  \tabcolsep=0.06cm
  \begin{tabular}{c c c c} 
    \includegraphics[width=0.245\linewidth]{color_comparison/000060_color.png} &
    \includegraphics[width=0.245\linewidth]{color_comparison/lidar.png}        &
    \includegraphics[width=0.245\linewidth]{color_comparison/gt.png}           &
    \includegraphics[width=0.245\linewidth]{color_comparison/ours.png} 
    \\
    (a) Input image & (b) Input lidar disparity & (c) Ground truth & (d) Ours
    \\
    \includegraphics[width=0.245\linewidth]{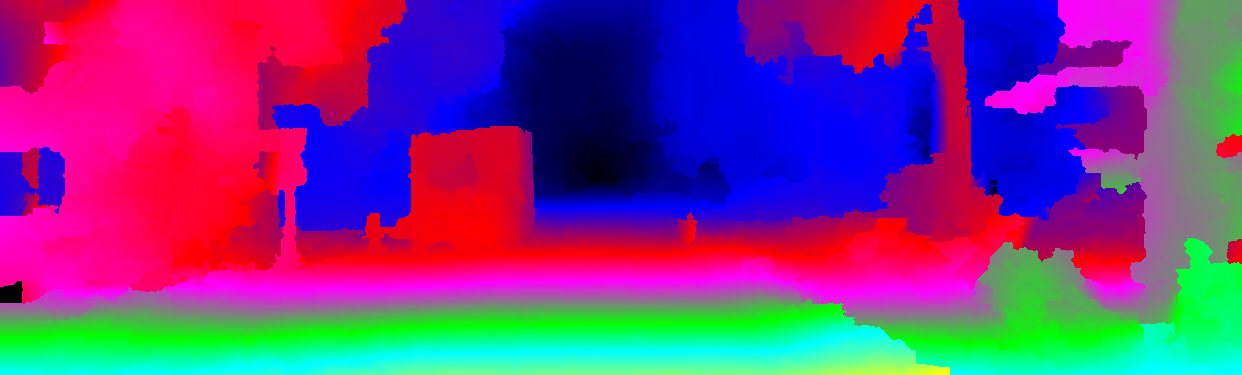}      &
    \includegraphics[width=0.245\linewidth]{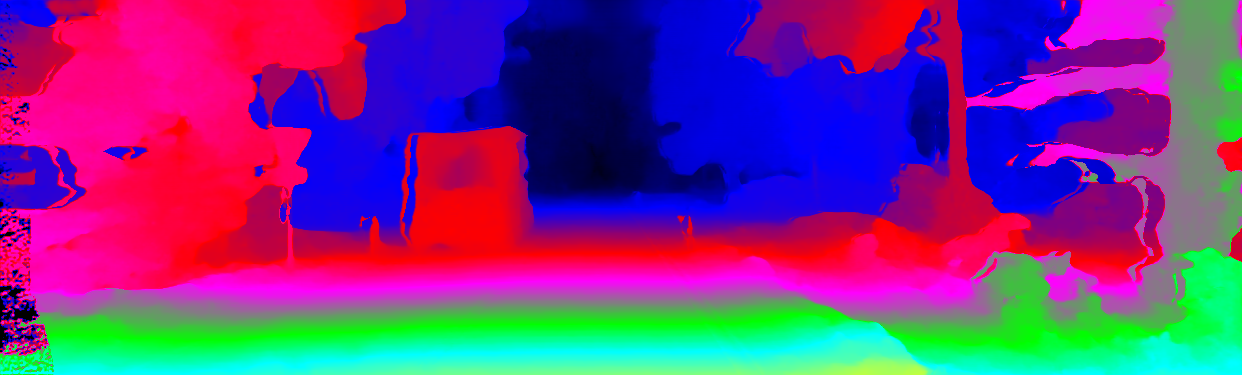}        &
    \includegraphics[width=0.245\linewidth]{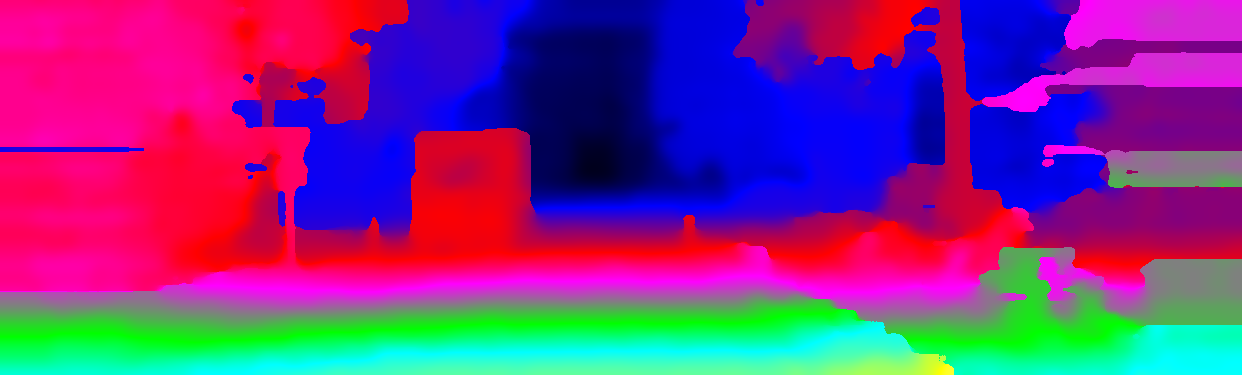} &
    \includegraphics[width=0.245\linewidth]{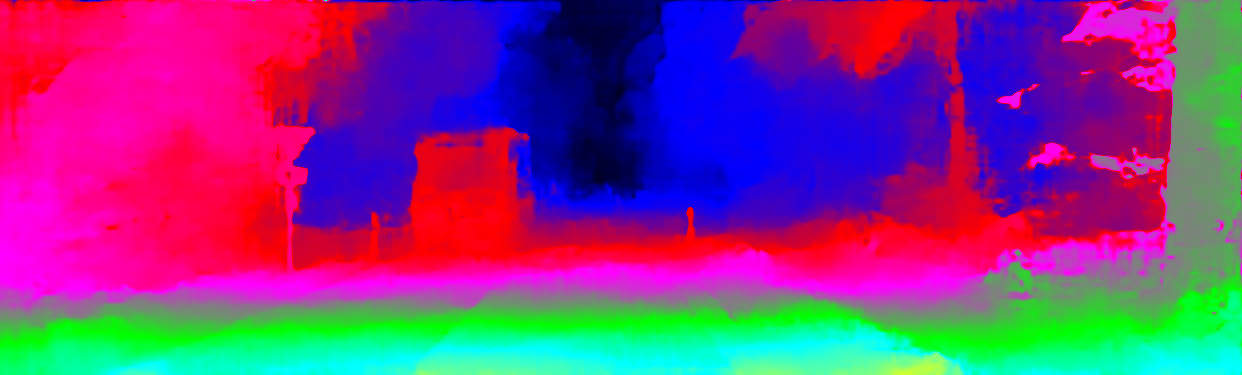}   
     \\
    (e)SPS-ST \cite{Yamaguchi14} & (f) SsSMNet \cite{zhong2017self} & (g) MC-CNN \cite{Zbontar2016} & (h) PSMnet \cite{chang2018pyramid} 
    \\    
  \end{tabular}  
  \caption{\textbf{Qualitative results of the methods from Table~\ref{tab:results_kitti141_add}.} Our method is trained on KITTI VO dataset and tested on the selected unseen KITTI 141 subset without any finetuning.} 
  \label{fig:results_kitti141_imgs} 
\end{figure*}

\section{Qualitative results on Synthia dataset}
In Fig.~\ref{fig:synthia_imgs}, we show qualitative comparison results on Synthia dataset. Our method achieves the lowest bad pixel ratio.

\end{document}